\documentclass[10pt,twocolumn,letterpaper]{article}

\usepackage[pagenumbers]{cvpr}

\usepackage{algorithmic}
\usepackage{array}
\usepackage{textcomp}
\usepackage{stfloats}
\usepackage{url}
\usepackage{graphicx}
\usepackage{amsmath,amssymb,amsfonts}
\usepackage{multirow,multicol}
\usepackage{todonotes}
\usepackage{microtype}
\usepackage{overpic}
\usepackage{booktabs}
\usepackage{colortbl}
\usepackage{bbm}

\definecolor{cvprblue}{rgb}{0.21,0.49,0.74}
\usepackage[
    pagebackref,
    breaklinks,
    colorlinks,
    allcolors=cvprblue
]{hyperref}

\title{Prior-free relative 6D pose estimation of multiple object instances}

\author{
Behdad Khodabandehloo$^{1,2}$ \qquad
Andrea Caraffa$^{1}$ \qquad
Davide Boscaini$^{1}$ \qquad
Fabio Poiesi$^{1}$\\[0.5em]
$^{1}$Fondazione Bruno Kessler, Trento, Italy \\
$^{2}$University of Trento, Trento, Italy
}

\begin{document}

\newcommand{\fabio}[1]{\todo[color=blue!20, inline, author=Fabio]{#1}}
\newcommand{\davide}[1]{\todo[color=yellow!20, inline, author=Davide]{#1}}
\newcommand{\andrea}[1]{\todo[color=green!20, inline, author=Andrea]{#1}}
\newcommand{\behdad}[1]{\todo[color=red!20, inline, author=Behdad]{#1}}

\newcommand{\warning}[1]{\textbf{\color{red!90}{#1}}}

\newcommand{\acronym}{PROSE\xspace}

\def\eg{\emph{e.g.}}
\def\ie{\emph{i.e.}}

\definecolor{myazure}{rgb}{0.8509,0.8980,0.9412}
\definecolor{mygreen}{RGB}{34,139,34}
\definecolor{myred}{RGB}{139,34,34}

\maketitle

\begin{abstract}
Object 6D pose estimation formulations have progressively reduced reliance on object-specific priors, evolving from explicit 3D models to multi-view object captures to single reference images. 
We take this progression to its extreme by introducing \emph{prior-free relative 6D pose estimation}, which lifts the assumption of knowing which object is to be posed within the scene. 
This novel setting aims to estimate the relative poses of multiple instances of an unknown object within the same image, without requiring CAD models, templates, or reference images. 
We solve this by formulating a novel method (PROSE) that finds coarse correspondences between object instances using multimodal foundation features, thus requiring no training. 
We refine these correspondences by imposing cycle consistency across tuples of instances, and leverage the resulting globally consistent correspondences to estimate the relative 6D pose between any pair of instances.
To enable systematic evaluation, we design a novel benchmark (PRENCH) built from three multi-instance BOP datasets and enriched with task-specific metadata.
PROSE consistently outperforms baselines obtained by adapting state-of-the-art single-image methods to the proposed setting, while requiring neither task-specific supervision nor additional learned components.
Project website: \url{https://tev-fbk.github.io/PROSE/}.
\end{abstract}


\vspace{-5pt}

\section{Introduction}\label{sec:intro}

Enabling robots to interact with objects in the physical world requires perception and reasoning~\cite{li2025developments, zheng2025survey}, with 6D object pose estimation being central for this~\cite{liu2026deep, nardon2025chip}.
While recent methods have achieved substantial progress in generalizing pose estimation to previously unseen objects, this generalization typically assumes that object-specific prior information is available at deployment time.

\begin{figure}[t!]
    \centering



    \begin{overpic}[trim=0 0 0 0, clip, width=1.0\linewidth]{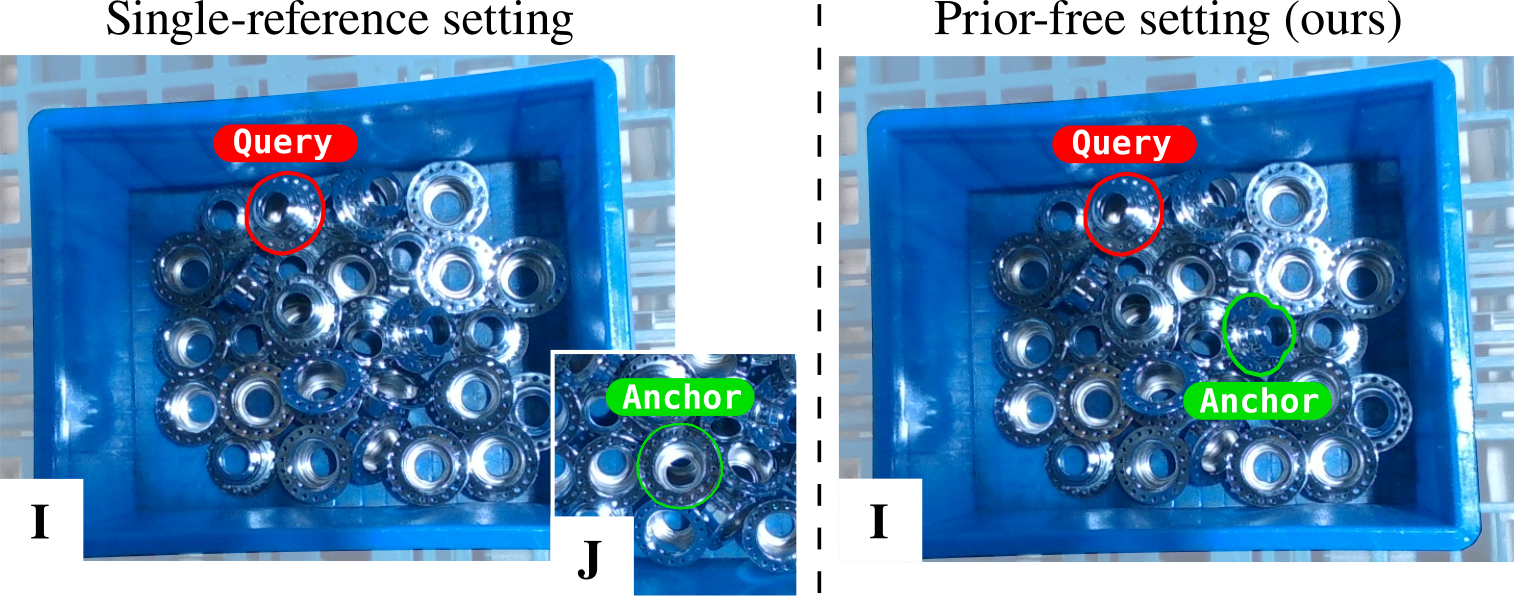}
    \end{overpic}

    \vspace{-2mm}
    \caption{
    We explore the novel task of \emph{prior-free relative 6D pose estimation} (right).
    Given an image $\mathbf{I}$ containing multiple instances of the same object, we estimate the 6D pose of a query instance (\textcolor{red}{$\bullet$}) relative to an anchor instance (\textcolor{green}{$\bullet$}).
    Unlike previous methods that require an external reference image $\mathbf{J}$ to specify the anchor (left), we select the anchor directly within the input image $\mathbf{I}$ (right).
    This removes the need for prior references, reducing setup time to zero and enabling practical deployment in real-world settings.
    }
    \label{fig:teaser}
\end{figure}

Existing approaches are progressively reducing the amount of such prior information.
\emph{Model-based} methods assume access to an explicit 3D representation of the target object, typically a CAD model or reconstructed mesh~\cite{labbe2022megapose, caraffa2024freeze, chen2024zeropose, nguyen2024gigapose, lin2024sam}; while highly effective, they depend on accurate object models that typically require dedicated reconstruction pipelines or manufacturer assets. 
\emph{Model-free multi-view} methods remove the need for an explicit 3D model, relying instead on a set of reference images or a video sequence capturing the target object from multiple viewpoints, collected beforehand~\cite{liu2022gen6d, sun2022onepose, park2020latentfusion}. 
\emph{Model-free single-reference} methods further reduce this requirement to a single image of the target object~\cite{liu2025one2any, kuang2026conceptpose, nguyen2024nope}. 
Despite this progression, all of these formulations retain a common assumption: some object-specific prior information, whether a 3D model, multiple views, or a single image, is provided before pose estimation can be performed.

This assumption can limit robotic applications where new objects must be introduced with minimal setup. 
For example, in high-mix low-volume manipulation, a robot may encounter a large variety of previously unseen objects while processing only a small number of each~\cite{chen2025zerobp, huang2025xyzibd, kleeberger2019large}: CAD models may be costly or not possible to obtain for every object, while collecting dedicated reference observations introduces an additional object-onboarding stage. 
Even when a single reference image is available, differences in viewpoint, occlusion, or visible surface area can limit the overlap between the reference image and the object observed during operation.
This led us to formulate the research question: \emph{How can we formulate object 6D pose estimation without any externally provided object reference at all?}

We address this question by introducing the novel problem formulation of prior-free relative 6D pose estimation of multiple object instances, considering a single RGBD observation containing multiple instances of the same unknown rigid object, with neither its CAD model nor any additional reference image assumed available. 
We exploit a simple observation: \emph{when several instances of the same object are simultaneously visible, the instances themselves can serve as mutual references}. 
One instance can be selected as an anchor pose, and the poses of the remaining instances can be estimated relative to it (Fig.~\ref{fig:teaser}). 
This setting removes object-specific acquisition before deployment while naturally matching scenarios such as industrial bin picking, where several instances of the same object coexist.  

We formulate a novel training-free method (\acronym\footnote{\underline{P}rior-free \underline{r}elative 6D p\underline{ose} estimation.}) that combines semantic~\cite{dinov2} and geometric~\cite{hamza2025dgedi} foundation features to establish correspondences between observed instances.
Unlike conventional relative-pose pipelines that process an anchor-query pair independently~\cite{kuang2026conceptpose, liu2025one2any}, \acronym further exploits the presence of multiple instances jointly, refining pairwise correspondences through cross-instance synchronization that enforces cycle consistency across all instances in the image. 
The resulting globally consistent correspondences are then used to estimate the relative 6D pose between any pair of instances through robust 3D registration.

We design a benchmark (PRENCH\footnote{\underline{P}rior-free \underline{r}elative 6D pose estimation b\underline{ench}mark.}) for prior-free relative 6D pose estimation constructed from the multi-instance BOP datasets IC-BIN~\cite{doumanoglou2016icbin}, IC-MI~\cite{tejani2014icmi}, and XYZ-IBD~\cite{huang2025xyzibd}.
PRENCH defines anchor-query relationships directly within each RGBD scene and enables evaluation across objects with substantially different appearance and geometric characteristics.
No existing method operates under this assumption, hence we adapt to the proposed setting One2Any~\cite{liu2025one2any} and ConceptPose~\cite{kuang2026conceptpose}, two related state-of-the-art single-reference (relative) pose estimation approaches.
Experiments on PRENCH show that \acronym consistently outperforms the adapted baselines across all three datasets.
In particular, \acronym outperforms our baselines by +19.2, +19.5, and +19.7 average recall on IC-BIN, IC-MI, and XYZ-IBD, respectively.
Compared with the best baseline for each dataset, it also reduces rotation and translation errors by 9.2--36.0 degrees and 7.7--15.3 mm, respectively.

\noindent In summary, our contributions are:
\begin{itemize}
    \item We present the novel task of prior-free relative 6D pose estimation, in which only the presence of multiple instances of the same object within the same image can be exploited to guide pose estimation.
    \item We propose \acronym, the first training-free method for this task, which explicitly leverages cycle consistency across multiple instances for accurate relative pose estimation.
    \item We introduce PRENCH, the first benchmark for prior-free relative 6D pose estimation, and establish strong baselines by adapting state-of-the-art single-image methods to the proposed setting.
\end{itemize}

\vspace{-2mm}

\section{Related works}\label{sec:related}

\noindent \textbf{Model-based methods}
Model-based methods require a 3D CAD model of the target object and generally follow either feature-matching or template-matching paradigms~\cite{liu2026deep,caraffa2025accurate}. They may be training-based, generalizing through large-scale training on diverse objects, or training-free, relying on frozen foundation or hand-crafted features. Among feature-matching approaches, FreeZe~\cite{caraffa2024freeze} fuses DINOv2 and GeDi descriptors for correspondence-based 3D registration, SAM-6D~\cite{lin2024sam} combines proposal selection with learned dense 3D--3D matching, and ZeroPose~\cite{chen2024zeropose} uses CAD-derived visual and geometric embeddings within a discovery--orientation--registration pipeline. Template-based approaches compare observation with rendered CAD views. MegaPose~\cite{labbe2022megapose} learns to score and refine rendered pose hypotheses, while GigaPose~\cite{nguyen2024gigapose} combines template retrieval with patch correspondences. FoundationPose~\cite{wen2024foundationpose} provides a learned framework for both model-based and model-free pose estimation. Despite their generalization to unseen objects, these methods still require an object-specific CAD model, whereas \acronym estimates poses directly from co-visible instances.

\noindent \textbf{Model-free multi-view methods} remove the need for a CAD model but require a video or multiple reference images of the target object. They broadly follow feature-matching or template-matching paradigms~\cite{liu2026deep}. Feature-matching methods establish 2D--3D correspondences using a reconstructed object representation or directly match features across the reference and query images. OnePose and OnePose++~\cite{sun2022onepose,he2022onepose++} reconstruct a 3D point model from reference views and match it to the query image using learned correspondences. Template-matching methods compare the query with available reference views or templates rendered from a reconstructed representation. Gen6D~\cite{liu2022gen6d} retrieves reference views in a learned feature space and subsequently refines the predicted pose. LatentFusion~\cite{park2020latentfusion} aggregates multiple observations into an implicit latent 3D representation used for pose estimation. Although these methods eliminate the need for a CAD model, they still require object-specific reference views.
\acronym uses co-visible instances from a single RGBD observation as mutual references and jointly refines their correspondences through cycle consistency.

\noindent\textbf{Model-free single-view methods} (also referred to as relative pose estimation methods) leverage information from a single reference image to estimate the pose of a query image relative to it.
These methods generally learn feature matching between two images to estimate the relative pose.
POPE~\cite{fan2024pope} is a training-free method and performs 6D pose estimation by leveraging DINOv2 features to directly perform feature matching between reference and query images.
H-/Oryon~\cite{corsetti2024oryon, corsetti2025horyon} are training-based methods that fuse DINO features with object-name text embeddings to perform segmentation and improve feature matching. 
NOPE~\cite{nguyen2024nope} directly learns to predict discriminative viewpoint embeddings that enable the relative pose of an object to be inferred in new images. 
One2Any~\cite{liu2025one2any} encodes a reference RGBD image into a pose-aware embedding that captures object geometry and appearance from a single view and decodes this representation to predict dense object-coordinate correspondences in the query image.
ConceptPose~\cite{kuang2026conceptpose} is a training-free method that leverages the explainability of vision–language models (VLMs) and enhances the extracted features using semantic concepts.
\acronym does not require additional views to serve as an anchor, as the anchor is defined as one of the object instances present in the query image. 

\vspace{-5pt}
\section{\acronym}\label{sec:method}

\begin{figure*}[ht!]
    \centering
    \includegraphics[width=0.9\linewidth]{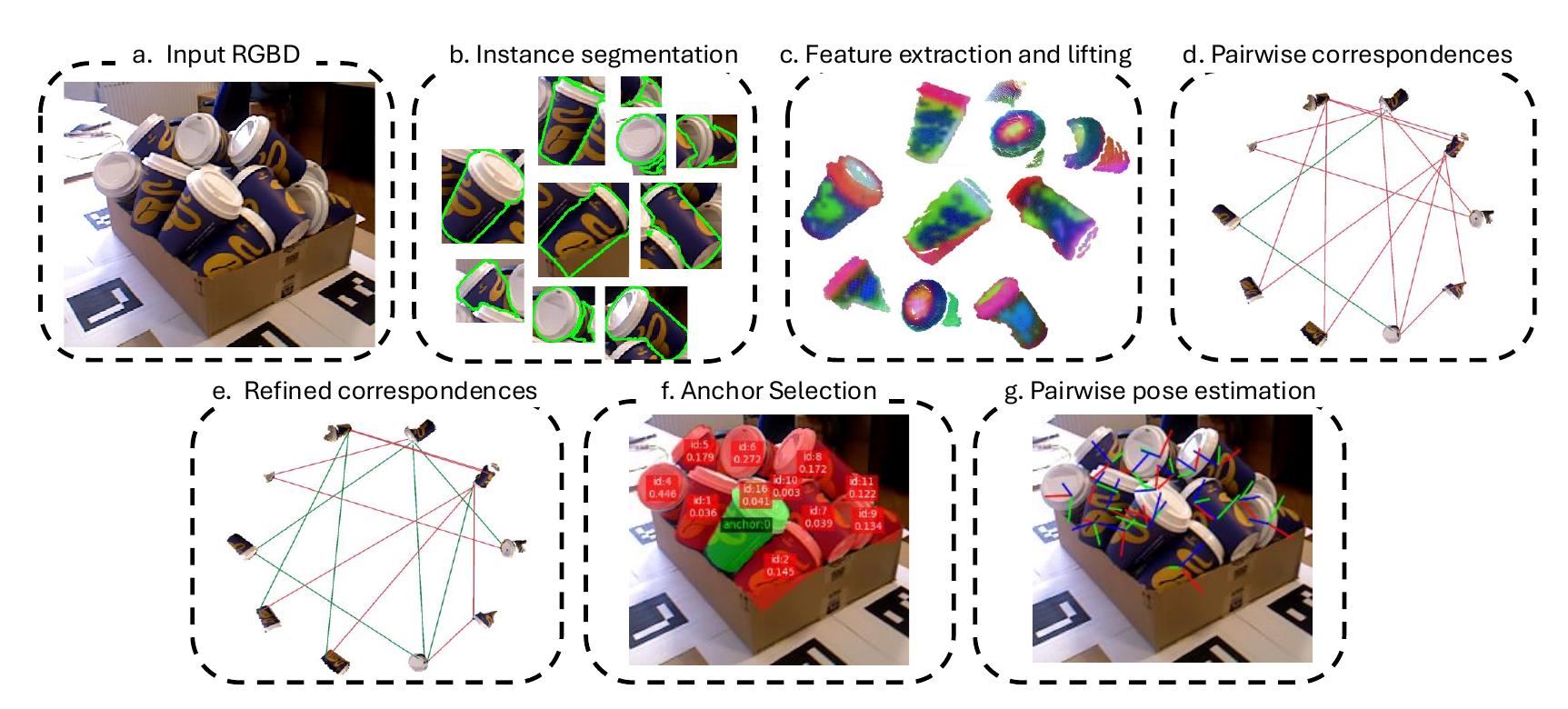}
    \vspace{-3mm}
    \caption{
    Overview of \acronym.
    Given an RGBD image containing multiple instances of the same object, our goal is to estimate their relative 6D poses without requiring any prior reference.
    (a) An RGBD image is provided as input.
    (b) Each instance is segmented. 
    (c) For each segmented instance, semantic and geometric features are extracted and fused after lifting the observations to 3D.
    (d) Initial pairwise correspondences are established via nearest-neighbor search in the feature space.
    (e) These correspondences are refined by enforcing cycle consistency across any instance tuples.
    (f) One instance is selected as the reference according to a predefined criterion (\eg, maximum visibility).
    (g) Finally, relative poses between this reference and all other instances are estimated using RANSAC-based 3D registration.
    }
    \label{fig:diagram}
\end{figure*}

\subsection{Overview}

\acronym consists of three steps, illustrated in Fig.~\ref{fig:diagram}.
Firstly, point-level features are extracted from the object instances using frozen foundation encoders and are fused together to obtain multimodal representations.
Secondly, reliable correspondences between point cloud pairs are estimated by solving a feature-based synchronization problem that enforces cycle consistency across tuples of instances.
Lastly, relative 6D poses between each pair of instances are computed by solving a RANSAC-based 3D registration problem.

\vspace{-2pt}
\subsection{Problem formulation}

We consider an RGBD image $\mathbf{I} \in [0, 255]^{H \times W \times 4}$ containing $N \ge 2$ instances $\{ \mathcal{O}_i, i=1, \dots, N \}$ of the same rigid object $\mathcal{O}$,
and assume that the intrinsic parameters $\mathbf{K} \in \mathbb{R}^{3\times3}$ of the sensor are known.
For each instance $\mathcal{O}_i$, we define $\mathbf{Y}_i \in \mathbb{R}^{V_i \times 2}$ as the set of its visible pixels in $\mathbf{I}$, and $\mathbf{X}_i \in \mathbb{R}^{V_i \times 3}$ as the point cloud obtained by lifting $\mathbf{Y}_i$ into 3D using $\mathbf{K}$ and depth measurements.
Our goal is to estimate, for every pair $i \neq j$, the relative rigid transformation $\mathbf{T}_{i,j} = (\mathbf{R}_{i,j} \in SO(3), \mathbf{t}_{i,j} \in \mathbb{R}^3)$ that best aligns $\mathbf{X}_i$ and $\mathbf{X}_j$, without using the 3D model of $\mathcal{O}$ or any other image as a reference.
We denote by $\mathbf{T}_{i,j}^\text{gt}$ the  ground-truth pose.

\subsection{Multimodal feature extraction}

For each object instance $\mathcal{O}_i$, we extract complementary appearance- and geometry-aware features and fuse them into a unified multimodal representation.
Specifically, we compute appearance-aware features by applying the visual encoder $\Phi_\Theta$ to the image crop associated with $\mathcal{O}_i$ to obtain $\mathbf{A}_i = \Phi_\Theta (\mathbf{Y}_i) \in \mathbb{R}^{V_i \times A}$, where $V_i$ denotes the number of visible pixels and $A$ is the dimensionality of the appearance feature space.
In parallel, we encode the corresponding point cloud $\mathbf{X}_i$ using the 3D encoder $\Psi_\Omega$ to obtain geometry-aware features $\mathbf{G}_i = \Psi_\Omega (\mathbf{X}_i) \in \mathbb{R}^{V_i \times G}$, where $G$ denotes the dimensionality of the geometry feature space.
The two modalities are subsequently fused by concatenating their features along the channel dimension, yielding the multimodal representation $\mathbf{F}_i = [ \mathbf{A}_i \mid \mathbf{G}_i ] \in \mathbb{R}^{V_i \times (A + G)}$ that jointly captures the visual appearance and 3D geometric structure of each object instance.
Finally, we perform row-wise L2 normalization on $\mathbf{F}_i$ to get the final features $\tilde{\mathbf{F}}_i$.

\vspace{-1mm}

\subsection{Coarse correspondence estimation}

For each pair $\mathbf{X}_i \in \mathbb{R}^{V_i \times 3}$, $\mathbf{X}_j \in \mathbb{R}^{V_j \times 3}$, we estimate a similarity map $\mathbf{S}_{i,j} \in [0, 1]^{V_i \times V_j}$ by computing the cosine similarity between the corresponding point features $\tilde{\mathbf{F}}_i$, $\tilde{\mathbf{F}}_j$, \ie, $ \mathbf{S}_{i,j} = \tilde{\mathbf{F}}_i \, \tilde{\mathbf{F}}_j^T $.
$\mathbf{S}_{i,j}$ can be interpreted as a soft correspondence map, from which we derive a discrete correspondence map $\Pi_{i,j}^0 \in \{0, 1\}^{V_i \times V_j}$ by replacing each non-empty row of $\mathbf{S}_{i,j}$ with a one-hot vector at its maximum.
The resulting map $\Pi_{i,j}^0$ provides an initial set of coarse correspondences between the two object instances.

\vspace{-1mm}
\subsection{Cross-instance correspondence refinement}

\begin{figure}[t!]
    \centering

    \begin{minipage}{0.48\linewidth}
    \centering
    \begin{overpic}[trim=550 0 0 0, clip, width=\linewidth]{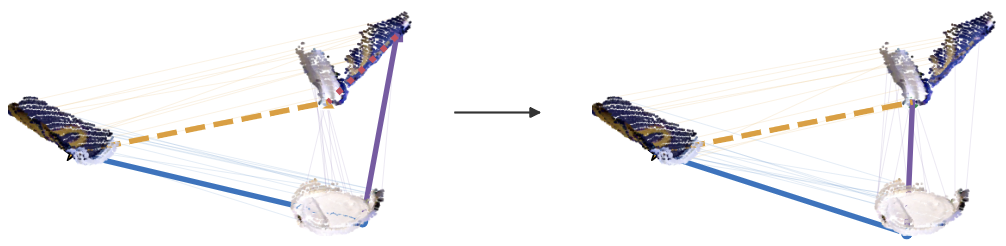}
        \put(82, -7){$\mathbf{X}_i$}
        \put(67, 50){$\mathbf{X}_j$}
        \put(10, 11){$\mathbf{X}_k$}
    \end{overpic}

    \vspace{-2mm}
    (a)
    \end{minipage}
    \hfill
    \begin{minipage}{0.48\linewidth}
    \centering
    \begin{overpic}[trim=0 0 550 0, clip, width=\linewidth]{main/figures/cycle_cons}
        \put(82, -7){$\mathbf{X}_i$}
        \put(67, 50){$\mathbf{X}_j$}
        \put(10, 10){$\mathbf{X}_k$}
    \end{overpic}

    \vspace{-2mm}
    (b)
    \end{minipage}

    \vspace{-1mm}
    \caption{
    Correspondences across three instances $\mathbf{X}_i$, $\mathbf{X}_j$, $\mathbf{X}_k$ of an IC-BIN coffee cup.
    (a) A consistent triplet: the purple, blue, and yellow correspondences satisfy cycle consistency, as composing the blue and yellow correspondences recovers the direct purple correspondence.
    (b) An inconsistent triplet: routing through $\mathbf{X}_k$ (blue + yellow) diverges from the direct correspondence (purple), producing the closure gap shown in red.
    }
    \label{fig:cycle_cons}
\end{figure}
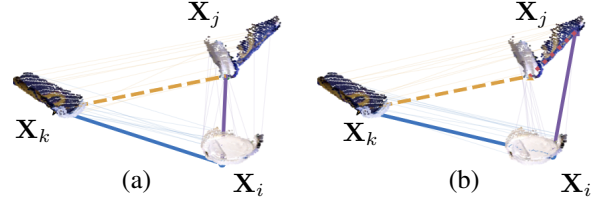

\noindent \textbf{Intuition.}
The coarse correspondences $\Pi_{i,j}^0$ are estimated independently for each pair of instances $\mathbf{X}_i$, $\mathbf{X}_j$, and may contain outliers when the corresponding features $\tilde{\mathbf{F}}_i$, $\tilde{\mathbf{F}}_j$ are not discriminative enough.
However, since all object instances depict the \emph{same} rigid object, we can leverage  correspondences between $\mathbf{X}_i$, $\mathbf{X}_j$ and other instances to refine $\Pi_{i,j}^0$ and filter out potential outliers.
Fig.~\ref{fig:cycle_cons} shows an example involving three instances $\mathbf{X}_i$, $\mathbf{X}_j$, $\mathbf{X}_k$.
Ideally, the direct correspondence from $\mathbf{X}_i$ to $\mathbf{X}_j$ should be consistent with the correspondence obtained by routing through $\mathbf{X}_k$, \ie 
\begin{equation}\label{eq:cycle_consistency}
\mathbf{C}_{i,j} = \mathbf{C}_{i,k} \mathbf{C}_{k,j}.
\end{equation}
This property is called \emph{cycle consistency}, and is illustrated in Fig.~\ref{fig:cycle_cons}(a).
Correspondences for which cycle consistency does not hold 
are considered outliers.
The deviation from the cycle consistency~\cite{huang2013consistent, pachauri2013solving}, \ie 
\begin{equation}\label{eq:closure_gap}
\lVert \mathbf{C}_{i,j} - \mathbf{C}_{i,k} \mathbf{C}_{k,j} \rVert_F,
\end{equation}
where $\lVert \cdot \rVert_F$ denotes the Frobenius norm, is referred to as the \emph{closure gap}  and illustrated by the red segment in Fig.~\ref{fig:cycle_cons}(b).
Our intuition is that coarse correspondences can be refined by minimizing the closure gap in Eq.~\ref{eq:closure_gap} across any tuple of instances.
Since multiple instances induce many overlapping cycles, enforcing this constraint jointly provides redundancy that suppresses isolated matching errors and produces a globally consistent set of correspondences.

\noindent \textbf{Optimization problem.}
Let $V=\sum_{i=1}^N V_i$ and $\mathbf{C} \in \mathbb{R}^{V \times V}$ denote the block-wise correspondence matrix obtained by stacking the pairwise correspondence matrices, \ie
\begin{equation}\label{eq:cmatrix}
\mathbf{C} = 
    \begin{bmatrix}
    \mathbf{I}_1 & \mathbf{C}_{1,2} & \dots & \mathbf{C}_{1,N} \\
    \mathbf{C}_{2,1} & \mathbf{I}_2 & \dots & \mathbf{C}_{2,N} \\
    \mathbf{C}_{N,1} & \mathbf{C}_{N,2} & \dots & \mathbf{I}_N
    \end{bmatrix},
\end{equation}
where $\mathbf{I}_i \in \mathbb{R}^{V_i \times V_i}$ is the identity matrix.
To enforce cycle consistency across all instances, one could minimize Eq.~\eqref{eq:closure_gap} over all triplets of instances.
However, this results in a combinatorial optimization problem that is NP-hard when the correspondence matrices are constrained to discrete assignments~\cite{shafique2005noniterative, pachauri2013solving, bernard2019hippi}.
Instead, we exploit an equivalent characterization: a family of pairwise maps is cycle-consistent if and only if it factors through a common set of $W$ latent object points~\cite{huang2013consistent, pachauri2013solving, zhou2015multi}, \ie \ there exist assignment matrices $\mathbf{U}_i \in \{0,1\}^{V_i \times W}$,
$\mathbf{U}_j \in \{0,1\}^{V_j \times W}$
with $\mathbf{C}_{i,j}=\mathbf{U}_i\mathbf{U}_j^{\top}$.
Equivalently, $\mathbf{C}=\mathbf{U}\mathbf{U}^{\top}$ with $\mathbf{U}=[\mathbf{U}_1^{\top},\dots,\mathbf{U}_N^{\top}]^{\top}$, so that $\mathbf{C}$ is positive semidefinite and $\operatorname{rank}(\mathbf{C})\le V'$.
Cycle consistency is thus equivalent to a low-rank constraint on $\mathbf{C}$.
By using nuclear norm as the convex relaxation of the hard constraint on the rank \cite{fazel2001rank, recht2010guaranteed, zhou2015multi}, we estimate $\mathbf{C}$ as the solution of the synchronization problem~\cite{huang2013consistent, zhou2015multi}
\begin{equation}\label{eq:sync}
\begin{aligned}
\min_{\mathbf{C}}\;&
-\langle \mathbf{S},\mathbf{C}\rangle
+\alpha\,\lVert\mathbf{C}\rVert_1
+\lambda\,\lVert\mathbf{C}\rVert_\ast \\
&\text{s.t.}\quad
\mathbf{0}\le\mathbf{C}\le\mathbf{1},\;\;
\mathbf{C}=\mathbf{C}^{\top}.
\end{aligned}
\end{equation}
where 
\begin{equation}
\mathbf{S} = 
    \begin{bmatrix}
    \mathbf{0}_1 & \mathbf{S}_{1,2} & \dots & \mathbf{S}_{1,N} \\
    \mathbf{S}_{2,1} & \mathbf{0}_2 & \dots & \mathbf{S}_{2,N} \\
    \mathbf{S}_{N,1} & \mathbf{S}_{N,2} & \dots & \mathbf{0}_N
    \end{bmatrix},
\end{equation}
$\langle \mathbf{S}, \mathbf{C} \rangle_F = \operatorname{Tr}( \mathbf{S}^{\top} \mathbf{C} )$ is the Frobenius inner product, $\operatorname{Tr}$ is the trace operator, $\lVert\mathbf{C}\rVert_1 = \sum_{a,b} |C_{ab}|$ is the entry-wise $\ell_1$ norm, $\lVert\mathbf{C}\rVert_\ast=\sum_{r}\sigma_r(\mathbf{C})$ is the nuclear norm (\ie, the sum of the singular values of $\mathbf{C}$). 
The data term $-\langle \mathbf{S},\mathbf{C}\rangle$ rewards assigning correspondence
weight to point pairs with high feature similarity, and it is balanced against two
regularizers weighted by the hyperparameters $\alpha,\lambda\in\mathbb{R}_{>0}$:
the term $\alpha\lVert\mathbf{C}\rVert_1$ promotes \emph{sparse} correspondences that
approximate a (partial) permutation matrix, assigning each point to a single match,
while $\lambda\lVert\mathbf{C}\rVert_\ast$ promotes the \emph{low-rank} structure that
enforces cycle consistency.

%
Let $\mathbf{C}^\ast \in [0, 1]^{V \times V}$ be the solution of the optimization problem~\eqref{eq:sync}.
We derive a discrete correspondence map $\Pi \in \{0, 1\}^{V \times V}$ by replacing each non-empty row of $\mathbf{C}^\ast$ with a one-hot vector at its maximum.
The final map $\Pi$ provides the set of refined correspondences between any object instances.

\noindent \textbf{Algorithm.}
Solving Eq.~\eqref{eq:sync} directly is infeasible for two reasons.
First, it requires computing the nuclear norm (and hence the SVD of a $V \times V$ matrix) at every iteration.
To avoid this, we use the variational nuclear norm~\cite{cabral2013unifying, zhou2015multi}
$\lVert\mathbf{C}\rVert_\ast=\min_{\mathbf{C}=\mathbf{A}\mathbf{B}^{\top}}
\tfrac12(\lVert\mathbf{A}\rVert_F^2+\lVert\mathbf{B}\rVert_F^2)$ and factorize
$\mathbf{C}=\mathbf{A}\mathbf{B}^{\top}$ with $\mathbf{A},\mathbf{B}\in\mathbb{R}^{V\times r}$, turning the penalty into a Frobenius regularization on the factors.

Second, solving Eq.~\eqref{eq:sync} at full point cloud resolution is intractable, since $\mathbf{C}$ is $V\times V$ with $V = \sum_i V_i$ in the tens of thousands. 
We therefore subsample each instance to $W=400$ points by farthest point sampling~\cite{qi2017pointnet++} ($V=NW$), and keep $\mathbf{S}$ sparse, filling only the nearest-neighbor matches between sampled points, weighted by cosine similarity and normalized to $[0,1]$.

\vspace{-1mm}
\subsection{Relative 6D pose estimation}

Given the refined correspondences $\Pi_{i,j}$ between $\mathbf{X}_i$ and
$\mathbf{X}_j$, we estimate their relative 6D pose $\mathbf{T}_{i,j}\in SE(3)$ via RANSAC-based 3D registration.
RANSAC is an iterative algorithm.
At each iteration $h$, it randomly samples a subset of three correspondences from $\mathbf{C}_{i,j}$ and estimates the associated rigid transformation $\mathbf{T}_{i,j}^h \in SE(3)$ in closed form via singular value decomposition (SVD).
Each hypothesis $\mathbf{T}_{i,j}^h$ is scored by the number of \emph{inlier correspondences}, \ie, correspondences for which $\lVert \mathbf{T}_{i,j}^h \mathbf{X}_i - \mathbf{X}_j \rVert_F < D_\text{RANSAC}$, where $D_\text{RANSAC}$ is a distance threshold hyperparameter.
This sampling loop is repeated for a fixed number of iterations $I_\text{RANSAC}$, after which the highest-scoring hypothesis is retained as the final transformation $\mathbf{T}_{i,j}$.
This procedure provides robustness to potential residual outliers in $\Pi_{i,j}$.

\vspace{-2mm}
\section{PRENCH}\label{sec:benchmark}

\vspace{-1mm} 
PRENCH is designed for assessing the performance of prior-free relative 6D pose estimation methods.
PRENCH is built upon data from three multi-instance BOP datasets, namely IC-BIN~\cite{doumanoglou2016icbin}, IC-MI~\cite{tejani2014icmi}, and XYZ-IBD~\cite{huang2025xyzibd}, and augments their original absolute 6D pose annotations with novel metadata tailored to the proposed prior-free relative pose estimation task.
In particular, we define a reference object instance, or \emph{anchor}, as the instance with the highest visibility for each image.
We also categorize the remaining instances into easy, medium, and hard cases according to their spatial overlap with the anchor, measured as intersection-over-union between the object 3D model and the partial instance observed in the image.
PRENCH comprises 1166 RGBD and grayscale-D images acquired with five different sensors: an Intel RealSense D415 stereo camera, a Photoneo PhoXi M 3D scanner, an XYZ Robotics AL-M DLP structured-light camera, a PrimeSense short-range RGBD sensor, and an Ensenso N20-1202-16-BL stereo camera.
This sensor diversity allows us to assess the robustness of relative pose estimation methods across substantially different sensing modalities and acquisition conditions.
PRENCH covers 21 distinct objects, ranging from everyday objects, such as coffee cups, juice boxes, and shampoo bottles, to industrial objects, including gears, wheels, and brackets.
While everyday objects often exhibit rich textures that can provide useful cues for resolving 6D pose ambiguities, industrial items frequently exhibit less informative appearances, such as uniform plastic or metallic surfaces, together with complex geometries.
This diversity provides a challenging testbed for assessing the generalization of relative pose estimation methods across both appearance and geometry.
The benchmark evaluates the anchor-to-instance relative pose estimation accuracy using average recall (AR)~\cite{hodavn2020bop} (the mean of $\mathrm{AR}_{\mathrm{VSD}}$, $\mathrm{AR}_{\mathrm{MSSD}}$, and $\mathrm{AR}_{\mathrm{MSPD}}$), ADD(S)~\cite{hinterstoisser2012model, xiang2017posecnn}, as well as rotation and translation errors (RE, TE).
In addition to reporting overall performance, PRENCH enables analysis across difficulty levels, object categories, and sensing conditions, providing a more detailed characterization of the strengths and limitations of prior-free relative 6D pose estimation approaches.
PRENCH will be publicly released upon acceptance.

\vspace{-6pt}
\section{Experiments}\label{sec:experiments}

\vspace{-1mm}
\subsection{Implementation details}

We use DINOv2-small~\cite{dinov2} as vision encoder $\Phi_\Theta$ with $A=384$-dimensional features and dGeDi~\cite{hamza2025dgedi} as geometric encoder $\Psi_\Omega$ with $G=64$-dimensional features. 
The resulting fused feature has dimension $F = A + G = 448$ after concatenation.
Eq.~\eqref{eq:sync}'s regularization parameters are set to $\lambda = 50$ and $\alpha = 0.2$.
We use the ADMM~\cite{boyd2011distributed} alternating least-squares optimizer, initializing $\mathbf{C}$ as $\mathbf{S}$ and running for a maximum of 40 iterations.
We estimate the rigid transformation between two object instances using the correspondence-based RANSAC implementation provided by Open3D~\cite{zhou1801open3d}, and set $D_\text{RANSAC} = 5$ $\text{mm}$ and $I_\text{RANSAC} = 100\text{k}$.

\vspace{-1mm}
\subsection{Comparison procedure}

We identified One2Any~\cite{liu2025one2any} and ConceptPose~\cite{kuang2026conceptpose} as the closest methods to our setting, as both estimate relative 6D pose from an external reference image depicting the object of interest.
We adapt both methods to our setting by constructing their required reference image directly from the input image, rather than relying on an externally provided one.
Following their evaluation procedure, we use ground-truth instance masks to crop objects. 
We crop the input image around the object instance identified as the anchor, and use this crop as the reference image.
All other instances are cropped accordingly and used to estimate the pose of the object within.
This adaptation preserves the reference-based formulation of both methods while removing dependence on a separately supplied reference view, allowing them to be evaluated within our prior-free setting.

\subsection{Quantitative results}

\renewcommand{\arraystretch}{0.9}
\begin{table*}[t]
\centering
\caption{
Quantitative results on PRENCH.
AR and ADD(S) are the higher the better ($\uparrow$).
RE and TE are the lower the better ($\downarrow$) and are measured in degrees and millimeters, respectively.
\textbf{Bold} denotes the best performance.
Results of our method are \colorbox{myazure}{highlighted}.
Row~4 shows our improvement over the second best method.
}
\label{tab:quant}
\vspace{-2mm}
\resizebox{\linewidth}{!}{%
\begin{tabular}{rl|cccc|cccc|cccc}
    \toprule
    & \multirow{2}{*}{Method} & \multicolumn{4}{c|}{IC-BIN} & \multicolumn{4}{c|}{IC-MI} & \multicolumn{4}{c}{XYZ-IBD} \\
    & & AR $\uparrow$ & ADD(S) $\uparrow$ & RE   $\downarrow$ & TE $\downarrow$ & AR $\uparrow$ & ADD(S) $\uparrow$ & RE $\downarrow$ & TE $\downarrow$ & AR $\uparrow$ & ADD(S) $\uparrow$ & RE $\downarrow$ & TE $\downarrow$ \\
    \toprule
    \color{gray} \footnotesize 1 & One2Any~\cite{liu2025one2any} & 17.7 & 61.0 & 70.5 & 26.9 & 47.7 & 98.0 & 47.2 & 12.3 & 9.2 & 49.2 & 85.0 & 33.2 \\
    \color{gray} \footnotesize 2 & ConceptPose~\cite{kuang2026conceptpose} & 18.8 & 58.8 & 76.2 & 36.6 & 50.8 & 80.2 & 51.1 & 29.1 & 36.1 & 70.1 & 64.6 & 24.4 \\
    \rowcolor{myazure} \color{gray} \footnotesize 3 & \acronym (ours) & \textbf{38.0} & \textbf{83.3} & \textbf{61.3} & \textbf{16.8} & \textbf{70.3} & \textbf{100} & \textbf{30.4} & \textbf{4.6} & \textbf{55.8} & \textbf{92.1} & \textbf{28.6} & \textbf{9.1} \\
    \midrule
    \color{gray} \footnotesize 4 & Improvement & \textcolor{mygreen}{+19.2} & \textcolor{mygreen}{+22.3} & \textcolor{mygreen}{-9.2} & \textcolor{mygreen}{-10.1} & \textcolor{mygreen}{+19.5} & \textcolor{mygreen}{+2.0} & \textcolor{mygreen}{-16.8} & \textcolor{mygreen}{-7.7} & \textcolor{mygreen}{+19.7} &
    \textcolor{mygreen}{+22.0} &
    \textcolor{mygreen}{-36.0} &
    \textcolor{mygreen}{-15.3} \\
    \bottomrule
\end{tabular}
}

\end{table*}

We evaluate \acronym's performance on the PRENCH benchmark and compare it against the results obtained with the revised One2Any and ConceptPose baselines.
Tab.~\ref{tab:quant} shows the results on the IC-BIN~\cite{doumanoglou2016icbin}, IC-MI~\cite{tejani2014icmi}, and XYZ-IBD~\cite{huang2025xyzibd} datasets in terms of AR, ADD(S), RE, and TE.
All metrics consider as valid also symmetric ground-truth poses generated using the object symmetries provided by the original BOP datasets metadata.
The datasets considered pose different challenges:
IC-BIN and IC-MI contain everyday objects, while XYZ-IBD contains industrial objects.
IC-BIN and XYZ-IBD contain several object instances randomly displaced in a bin with no other objects present, making self-occlusion the main challenge.
In contrast, IC-MI contains far fewer instances (three on average), but these are randomly positioned on a table where other objects and clutter act as distractors.
\acronym outperforms both competitors across all three datasets by a significant margin on all metrics.
AR improves by about 20 points across all datasets, despite being a very strict metric (entries on the BOP public leaderboard typically differ only by decimal variations in AR).
ADD(S) has a similar improvement of +22 points in IC-BIN and XYZ-IBD, with a more modest improvement on IC-MI.
Rotation estimation is, on average, 9.2--36.0 degrees more accurate, while translation estimation improves by 7.7--15.3 mm.
The improvement is largest in the more challenging setting provided by XYZ-IBD, with real-world industrial scenes and metallic mechanical components.
There is no clear ranking among the competitors: One2Any typically yields lower rotation and translation errors, while ConceptPose achieves better AR and ADD(S) on average.

\vspace{-3pt}
\subsection{Qualitative results}

\begin{figure*}[t!]
    \centering
    \begin{minipage}{\textwidth}

    \centering 
    \begin{minipage}{0.245\textwidth}
    \centering
    Input image
    \end{minipage}
    \hfill
    \begin{minipage}{0.245\textwidth}
    \centering
    One2Any~\cite{liu2025one2any}
    \end{minipage}
    \hfill
    \begin{minipage}{0.245\textwidth}
    \centering
    ConceptPose~\cite{kuang2026conceptpose}
    
    \end{minipage}
    \hfill
    \begin{minipage}{0.245\textwidth}
    \centering
    \acronym (ours)
    \end{minipage}

    \vspace{0.5mm}
    \begin{minipage}{\textwidth}
    \centering
    \begin{overpic}[trim=10 10 0 10, clip, width=\linewidth]{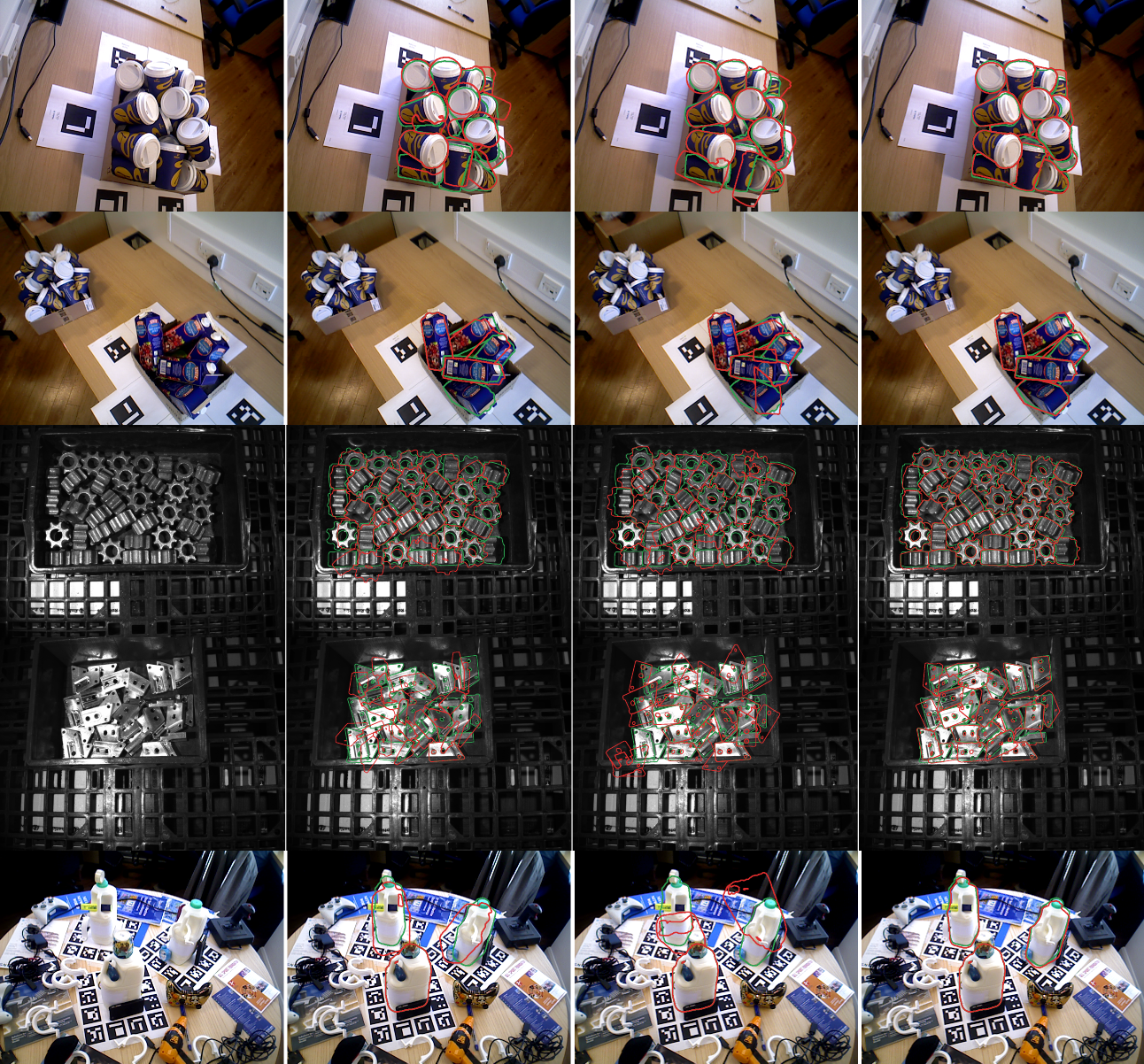}
    \end{overpic}
    \end{minipage}
    \end{minipage}

    \vspace{-1mm}
    \caption{
    Qualitative results on PRENCH.
    Each row shows a different scene.
    Green borders denote the ground-truth 6D poses, while red borders show each method's predictions.
    Accurate poses show good alignment between the two colored borders.
    From top to bottom: IC-BIN coffee cups (row 1) and juice boxes (row 2), XYZ-IBD metallic gears (row 3) and brackets (row 4), and IC-MI milk bottles (row 5).
    Each column shows a different method.
    From left to right: the RGB channels of the input image (column 1), and the predictions of One2Any (column 2), ConceptPose (column 3), and \acronym (column 4).
    \acronym is more accurate than competitors.
    }
    \label{fig:qual}
\end{figure*}

Fig.~\ref{fig:qual} shows some qualitative results on PRENCH.
Rows show different scenes: IC-BIN coffee cups (row 1) and juice boxes (row 2), and XYZ-IBD metallic gears (row 3) and brackets (row 4).
The left-most column shows the input images, while the remaining ones compare the predictions of \acronym (column 2), One2Any (column 3), and ConceptPose (column 4).
Green borders denote the ground-truth 6D poses, while red borders show each method's predictions.
The degree of alignment between the two colored borders indicates pose accuracy: \acronym is significantly more accurate than its competitors.
In the IC-BIN scenes (rows~1--2), self-occlusion among densely packed cups and boxes often causes One2Any and ConceptPose to predict wrong poses, while PROSE's cross-instance refinement can recover correct correspondences, thus producing correct pose estimates. 
In the XYZ-IBD scenes (rows~3--4), textureless metallic gears and brackets provide little appearance signal, and both baselines produce scattered, inaccurate predictions, while PROSE instead remains accurate. 
In the IC-MI scene (row~5), similarly to the two previous cases, objects have little texture information, and we can see how PROSE can still correctly pose the objects, unlike One2Any and ConceptPose.

\vspace{-4pt}
\subsection{Ablation study}

Tab.~\ref{tab:ablation_features} evaluates a variation of our pipeline in which we deactivate the cross-instance synchronization module and directly perform RANSAC-based relative pose estimation from the multimodal foundation features (\ie, the input to the pose estimation module is $\Pi^0$ instead of $\Pi$). This ablation uses full point clouds, and results are reported in terms of AR.
Row~1 reports results using only DINOv2~\cite{dinov2} visual features aware of both textures and semantics, row~2 using only dGeDi~\cite{hamza2025dgedi} geometric features, and row~3 when using their concatenation. 
The two datasets exhibit complementary characteristics:
visual features shine on IC-BIN, which contains everyday objects that are generally rich in texture,
whereas geometric features shine on XYZ-IBD, which consists of metallic industrial objects, where the object shapes are the primary cues to guide pose estimation.
Combining DINOv2 and dGeDi features consistently improves the pose accuracy across the two datasets, highlighting the benefit of leveraging complementary semantic and geometric representations.
These results outperform those obtained by One2Any~\cite{liu2025one2any} and ConceptPose~\cite{kuang2026conceptpose} in Tab.~\ref{tab:quant}, showing that frozen foundation features alone can already provide stronger performance than both a method supervised on 6D pose estimation-specific data (One2Any) and a VLM-based method (ConceptPose). 

\renewcommand{\arraystretch}{0.9}
\begin{table}[t!]
\centering
\tabcolsep 9pt
\caption{
Contribution of the feature extraction module to final pose accuracy (AR) when no correspondence refinement is applied and pose estimation is performed on full point clouds, \ie, when $\Pi^0$ is fed to RANSAC-based pose estimation instead of $\Pi$.
Our default setting is \colorbox{myazure}{highlighted}.
}
\label{tab:ablation_features}
\vspace{-2mm}
\resizebox{\linewidth}{!}{%
\begin{tabular}{rlccc}
    \toprule
    & Features & IC-BIN & IC-MI & XYZ-IBD\\
    \toprule
    \color{gray} \footnotesize 1 & Appearance only & 36.2 & 69.8  & 36.3 \\
    \color{gray} \footnotesize 2 & Geometry only & 30.7 & 59.5 & 46.8 \\
    \rowcolor{myazure} \color{gray} \footnotesize 3 & Fused & \textbf{37.8} & \textbf{71.8} & \textbf{48.5} \\
    \bottomrule
\end{tabular}
}
\end{table}

Tab.~\ref{tab:ablation_refinement} evaluates the contribution of the synchronization module, which refines the coarse correspondences obtained from multimodal foundation features by imposing cycle consistency across object instances.
We report the rotation and translation errors for each scene of XYZ-IBD before (left column) and after (central column) refinement, as well as the absolute improvement (right column).
Row 16 shows the average rotation and translation errors across scenes (which differs from the average across instances reported in Tab.~\ref{tab:quant}).
\acronym predicts more accurate poses, both in rotation (-3.2 degrees on average) and translation (-1.8 mm on average) compared to the refinement-free baseline obtained by feeding the RANSAC-based relative pose estimation module with $\Pi^0$ instead of $\Pi$.
The gain is consistent throughout all scenes, and more pronounced for the more challenging scenes (rows 2, 9, and 13).

\newcolumntype{H}{>{\columncolor{myazure}}r}
\renewcommand{\arraystretch}{0.9}
\begin{table}[t!]
\centering
\tabcolsep 9pt
\caption{
Contribution of the correspondence refinement module design to final pose accuracy on XYZ-IBD scenes before and after applying the refinement procedure, and measured in terms of RE and TE.
Our default setting is \colorbox{myazure}{highlighted}.
Imp. denotes the absolute improvement.
Row 16 shows the average across scenes, which differs from the average across instances reported in Tab.~\ref{tab:quant}.
}
\label{tab:ablation_refinement}
\vspace{-2mm}
\resizebox{\linewidth}{!}{%
\begin{tabular}{rr|rHr|rHr}
    \toprule
    & \multirow{2}{*}{Scene} & \multicolumn{3}{c|}{RE (degrees) $\downarrow$} & \multicolumn{3}{c}{TE (mm) $\downarrow$} \\
    & & Before & After & Imp. & Before & After & Imp. \\
    \toprule
    \color{gray} \footnotesize 1 & 0 & 6.8 & \textbf{5.8} & \textcolor{mygreen}{-1.0} & 2.8 & \textbf{2.5} & \textcolor{mygreen}{-0.3} \\
    \color{gray} \footnotesize  2 & 5 & 51.8 & \textbf{43.2} & \textcolor{mygreen}{-8.6} & 23.0 & \textbf{15.8} & \textcolor{mygreen}{-7.2} \\
    \color{gray} \footnotesize  3 & 10 &  9.0 & \textbf{ 8.1} & \textcolor{mygreen}{-0.9} &  4.9 & \textbf{ 4.7} & \textcolor{mygreen}{-0.2} \\
    \color{gray} \footnotesize  4 & 15 & 22.0 & \textbf{19.4} & \textcolor{mygreen}{-2.6} &  7.9 & \textbf{ 7.2} & \textcolor{mygreen}{-0.7} \\
    \color{gray} \footnotesize  5 & 20 & 24.3 & \textbf{22.3} & \textcolor{mygreen}{-2.0} &  9.0 & \textbf{ 8.0} & \textcolor{mygreen}{-1.0} \\
    \color{gray} \footnotesize  6 & 25 & 12.5 & \textbf{ 6.7} & \textcolor{mygreen}{-5.8} &  6.2 & \textbf{ 5.4} & \textcolor{mygreen}{-0.8} \\
    \color{gray} \footnotesize  7 & 30 & 36.8 & \textbf{33.4} & \textcolor{mygreen}{-3.4} & 10.7 & \textbf{ 8.9} & \textcolor{mygreen}{-1.8} \\
    \color{gray} \footnotesize  8 & 35 &  3.8 & \textbf{ 3.6} & \textcolor{mygreen}{-0.2} &  6.6 & \textbf{ 6.4} & \textcolor{mygreen}{-0.4} \\
    \color{gray} \footnotesize  9 & 40 & 19.3 & \textbf{12.0} & \textcolor{mygreen}{-7.3} &  6.1 & \textbf{ 5.3} & \textcolor{mygreen}{-0.8} \\
    \color{gray} \footnotesize 10 & 45 & 69.9 & 69.9 & 0.0 & 10.7 & \textbf{10.5} & \textcolor{mygreen}{-0.2} \\
    \color{gray} \footnotesize 11 & 50 & 53.6 & \textbf{51.1} & \textcolor{mygreen}{-2.5} & 16.3 & \textbf{15.0} & \textcolor{mygreen}{-1.3} \\
    \color{gray} \footnotesize 12 & 55 & 74.2 & \textbf{69.4} & \textcolor{mygreen}{-4.8} & 32.8 & \textbf{24.5} & \textcolor{mygreen}{-8.3} \\
    \color{gray} \footnotesize 13 & 60 & 24.1 & \textbf{15.7} & \textcolor{mygreen}{-8.4} & 12.7 & \textbf{10.6} & \textcolor{mygreen}{-2.1} \\
    \color{gray} \footnotesize 14 & 65 & \textbf{28.3} & 30.2 & \textcolor{myred}{+1.9} & 23.0 & \textbf{21.9} & \textcolor{mygreen}{-1.1} \\
    \color{gray} \footnotesize 15 & 70 & 87.8 & \textbf{85.5} & \textcolor{mygreen}{-2.3} & 13.7 & \textbf{12.6} & \textcolor{mygreen}{-1.1} \\
    \midrule
    \color{gray} \footnotesize 16 & Avg & 34.9 & \textbf{31.7} & \textcolor{mygreen}{-3.2} & 12.4 & \textbf{10.6} & \textcolor{mygreen}{-1.8} \\
    \bottomrule
\end{tabular}
}
\end{table}

\begin{figure}[t!]
    \centering
    \includegraphics[width=\columnwidth]{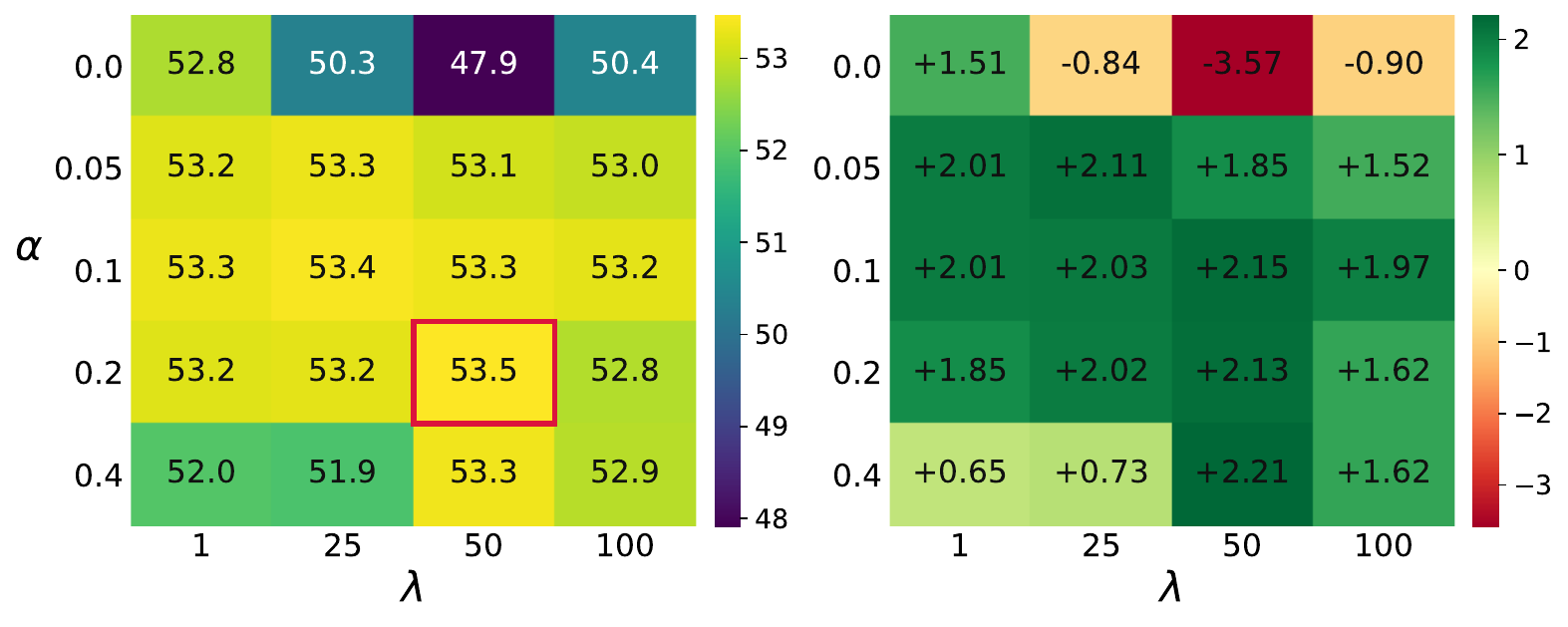}
    \vspace{-8mm}
    \caption{
    Hyperparameter ablation of the refinement on XYZ-IBD. 
    AR (\%) after refinement (left; best $\alpha{=}0.2$, $\lambda{=}50$ boxed) and the gain over
    the coarse baseline,
    $\Delta$AR $=$ after $-$ before (right). 
    Refinement is
    robust to $\lambda$ but needs a non-zero sparsity weight $\alpha$: $\alpha{=}0$ can
    reverse the gain ($-3.57$ at $\lambda{=}50$), while any $\alpha{>}0$ gives a consistent improvement of about 2 points.
    }
    \label{fig:alpha_lambda}
\end{figure}

\cref{fig:alpha_lambda} shows an analysis for $\alpha$ and $\lambda$ on XYZ-IBD (scenes with asymmetric objects)  and we find that performance is largely insensitive to $\lambda$, with any $\alpha>0$ yielding a consistent ${\sim}{+}2$
point gain over the coarse correspondences. 
The sparsity weight $\alpha$ is key. 
At  $\alpha=0$, the refinement is not effective, degrading the coarse baseline by $3.6$ points at $\lambda=50$. 
A small $\alpha$ already recovers the full benefit, forming a stable plateau across $\alpha\in[0.05,0.4]$; we adopt $\alpha=0.2$, $\lambda=50$.

\begin{figure}[t!]
    \centering
    \includegraphics[width=\columnwidth]{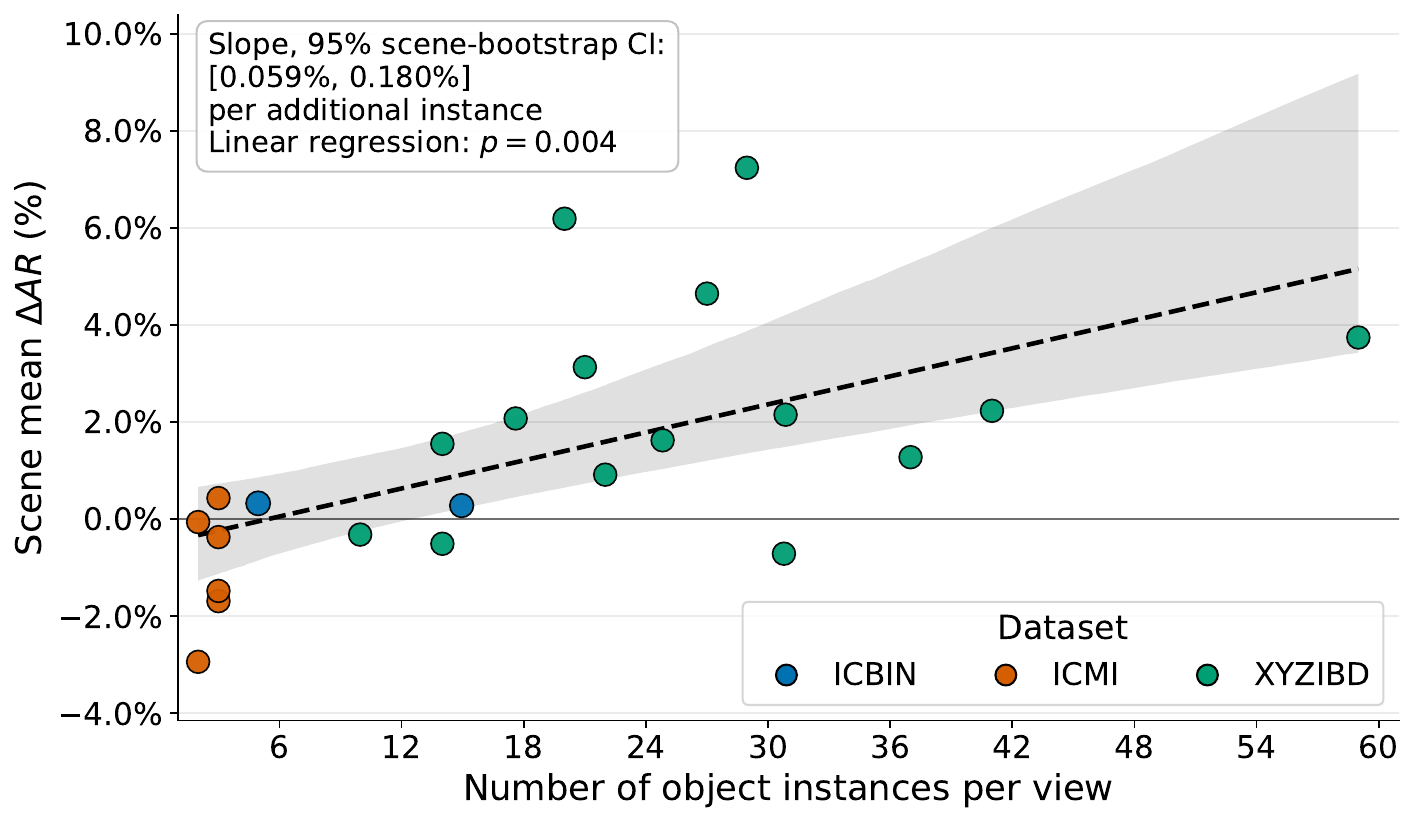}

    \vspace{-2mm}
    \caption{
    Relative AR improvement over the refinement-free baseline (y-axis) as a function of the number of object instances (x-axis). 
    The positive correlation indicates that the contribution of our refinement stage increases as more object instances become available.
    }
    \label{fig:ins_num_vs_AR}
\end{figure}

Fig.~\ref{fig:ins_num_vs_AR} shows the AR improvement of \acronym over the refinement-free baseline as a function of the number of object instances present in the input image.
The resulting distribution shows a positive correlation (dashed black curve), indicating that the contribution of \acronym's refinement stage increases as more object instances become available.
A similar trend emerges when comparing the improvements on the other metrics, namely ADD(S), RE, and TE.
The presence of many instances makes the relative pose estimation task more challenging, yet it is in this regime that our refinement stage proves most beneficial.

\vspace{-6pt}

\section{Conclusion}\label{sec:conclusion}

{\looseness=-1
We presented the novel task of prior-free relative 6D pose estimation of multiple object instances.
By eliminating the need to process externally provided object references (\eg, images, 3D models), our setting eliminates the overhead required by existing approaches, with practical applications in robotic manipulation.
We also presented \acronym, the first training-free method specifically designed for this setting.
\acronym explicitly exploits the complementary information available across multiple object instances present in the input image to refine the coarse correspondences estimated from multimodal foundation features.
Then, we introduced PRENCH, the first benchmark for prior-free relative 6D pose estimation.
PRENCH extends existing multi-instance BOP datasets~\cite{doumanoglou2016icbin, tejani2014icmi, huang2025xyzibd} with task-specific metadata extracted from the original absolute 6D pose annotations, providing a standardized framework for evaluating methods in this setting.
Experiments demonstrate that \acronym substantially outperforms state-of-the-art relative 6D pose estimation methods that rely on a single external reference image~\cite{liu2025one2any, kuang2026conceptpose}, when adapted to fit our setting.
\acronym assumes that instance masks and a predefined anchor are available. 
Although the refinement combines semantic and geometric features, it enforces feature and cycle consistency without explicitly considering the rigid geometric structure of correspondences. 
Future work can investigate automatic instance segmentation and geometry-aware synchronization that incorporates constraints such as pairwise-distance preservation to reject feature-consistent but geometrically incompatible correspondences.}

{
    \small
    \bibliographystyle{ieeenat_fullname}
    \bibliography{main}
}

\clearpage
\appendix

\twocolumn[
\begin{center}
    \vspace{-1em}
    {\LARGE\bfseries Supplementary material\par}
    \vspace{0.5em}
    {\large Prior-free relative 6D pose estimation of multiple object instances\par}
    \vspace{1.5em}
\end{center}
]

\section{Overview}

This supplementary material provides additional analyses, quantitative results, and qualitative visualizations that complement the experiments reported in the main paper.
In Sec.~\ref{sec:PRENCH_Stat}, we provide detailed statistics of the PRENCH benchmark, including the number of images, object instances, anchor-instance pairs, and objects in each dataset, as well as the distribution of instances per view and anchor-instance overlap.
In Sec.~\ref{sec:extended_ablation}, we present extended ablation studies that further analyze the behavior of \acronym under different levels of anchor-instance overlap. We then investigate the effect of cycle consistency refinement on each scene.
These experiments provide a more detailed view of the impact of global correspondence refinement via cycle-consistency in the accuracy of predicted poses.
In Sec.~\ref{sec:additional_quant}, we provide additional quantitative results for the correspondence refinement stage, including both per-scene and dataset-level comparisons before and after refinement.
These results complement the evaluation in the main paper and further validate the effect of multi-instance reasoning across IC-BIN~\cite{doumanoglou2016icbin}, IC-MI~\cite{tejani2014icmi}, and XYZ-IBD~\cite{huang2025xyzibd}.
Finally, in Sec.~\ref{sec:additional_qual}, we present additional qualitative results illustrating the behavior of the semantic (DINOv2~\cite{dinov2}), geometric (dGeDi~\cite{hamza2025dgedi}), and fused feature representations.
We visualize both the learned feature representations and the resulting inlier correspondences across textured everyday objects and textureless industrial objects.
We additionally visualize individual cycle-consistency refinement examples, including both successful and failure cases, to clarify how cycle-consistency refinement modifies the initial pairwise correspondences.

\section{Benchmark statistics}\label{sec:PRENCH_Stat}

\begin{figure*}[!htbp]
    \centering
    \includegraphics[width=1.0\linewidth]{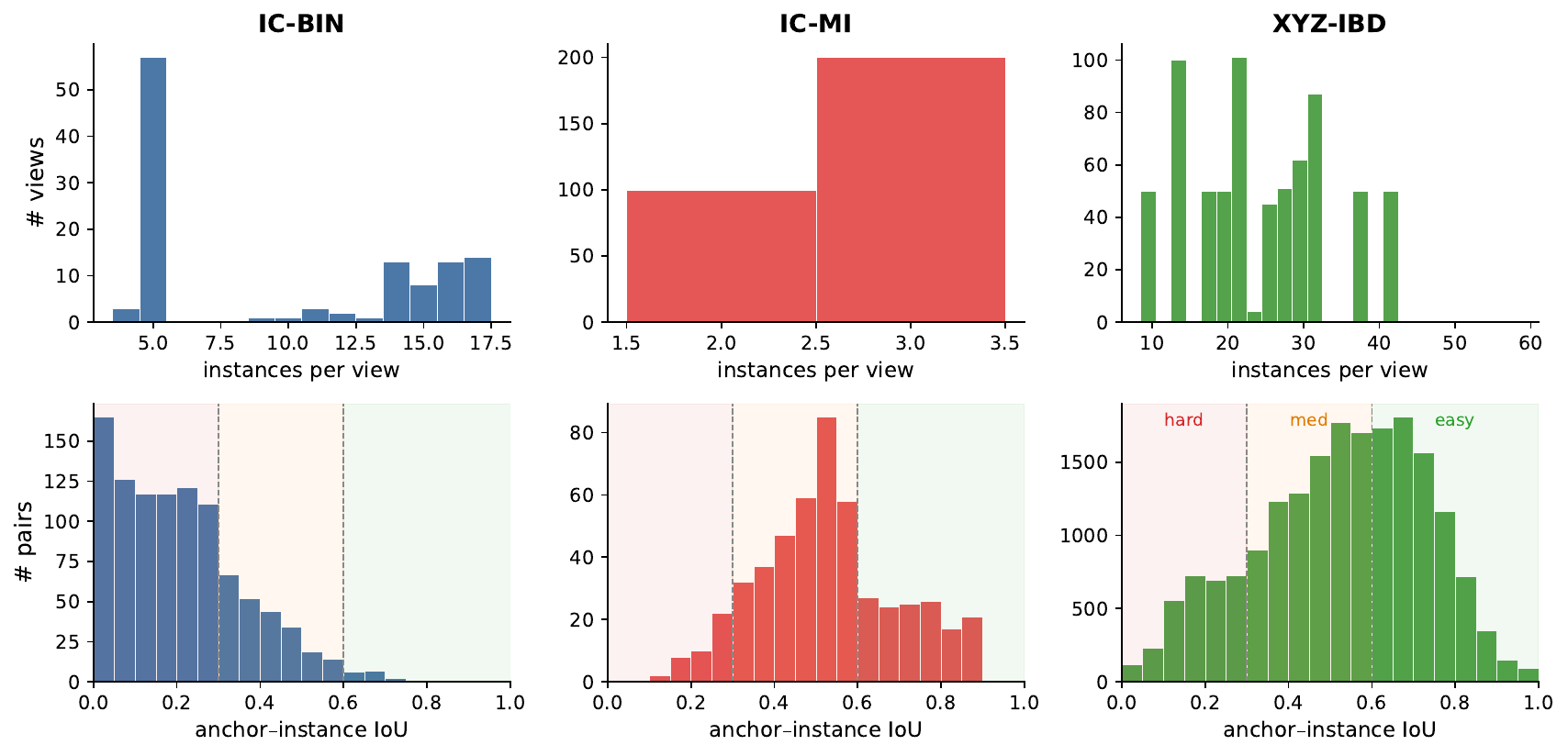}

    \caption{Top row: number of instances per view for
    each dataset. Bottom row: distribution of anchor–instance IoU, with the
    easy/medium/hard difficulty bands shaded (thresholds at $0.3$ and $0.6$). XYZ-IBD and
    IC-BIN contain many instances per image (mean $26.5$ and $9.8$, up to $59$ and $17$),
    whereas IC-MI has only ${\sim}3$; correspondingly, IC-BIN is skewed toward low overlap
    (hard), IC-MI concentrates in the medium range, and XYZ-IBD spreads across all
    difficulties.}
    
    \label{fig:prench_dist}
\end{figure*}

Fig.~\ref{fig:prench_dist} and Tab.~\ref{tab:prench_stats} summarize PRENCH.
In total the benchmark comprises $1{,}166$ images and $21{,}833$ object instances across three datasets and $21$ distinct objects, giving $20{,}667$ anchor–instance pairs for evaluation.
The datasets cover different number of per-view instances: XYZ-IBD averages $26.5$ instances per view (up to $59$), IC-BIN $9.8$, and IC-MI only $2.7$ (Fig.~\ref{fig:prench_dist}, top).
This diversity is central to our analysis, as the cross-instance refinement exploits redundant correspondences; consistent with this, its benefit is largest on the dense XYZ-IBD scenes and negligible on the sparse IC-MI ones (cf.\ Fig.~\ref{fig:delta_bars} and the per-scene analysis in Tab.~\ref{tab:refine_perscene}).
The datasets also differ in difficulty, quantified by the anchor–instance overlap, splitting instances into easy (IoU\,$>$\,0.6), medium (0.3\,$<$\,IoU\,$<$\,0.6), and hard
(0\,$<$\,IoU\,$<$\,0.3)(Fig.~\ref{fig:prench_dist}, bottom). IC-BIN mainly contains low-overlap (hard) pairs., IC-MI contains medium range overlap, and XYZ-IBD is spread across the full range.
Combined with a diverse object set (everyday textured objects in IC-BIN and IC-MI versus textureless, industrial parts in XYZ-IBD), PRENCH forms a broad evaluation setting that emphasizes relative pose estimation under both varying instance counts and varying degrees of overlap and occlusion.

\begin{table*}[t]
\centering
\tabcolsep 10pt
\caption{
PRENCH statistics across individual datasets (rows 1--3) and overall (row 4).
Columns show, from left to right, the number of objects, the number of images, the number of object instances (with details about per-image average and maximum), and the number of anchor-instance pairs (with details about difficulty levels: easy for anchor-instance pairs with IoU$>$0.6, medium with $0.3\le$IoU$\le$0.6, and hard with IoU$<$0.3).
$^\ast$Two objects recur across IC-BIN and IC-MI, so the total number of distinct objects is 21 instead of 23.
}
\label{tab:prench_stats}

\vspace{-2mm}
\resizebox{\linewidth}{!}{%
\begin{tabular}{rl|rr|rrr|rrrr}
    \toprule
    & \multirow{2}{*}{Dataset} & \multirow{2}{*}{\#Obj} & \multirow{2}{*}{\#Img} & \multicolumn{3}{c|}{\#Instances} & \multicolumn{4}{c}{\#Pairs} \\
    & & & & Total & Avg & Max & Total & Easy & Medium & Hard \\
    \midrule
    \color{gray} \footnotesize 1 & IC-BIN & 2 & 116 & 1{,}134 & 9.8 & 17 & 1{,}018 & 15 & 230 & 757 \\
    \color{gray} \footnotesize 2 & IC-MI & 6 & 300 & 800 & 2.7 & 3 & 500 & 140 & 318 & 42 \\
    \color{gray} \footnotesize 3 & XYZ-IBD & 15 & 750 & 19{,}899 & 26.5 & 59 & 19{,}149 & 7{,}592 & 8{,}445 & 3{,}047 \\
    \midrule
    \rowcolor{myazure} \color{gray} \footnotesize 4 & PRENCH & 21$^\ast$ & 1{,}166 & 21{,}833 & 13.0 & 59 & 20{,}667 &7{,}747 & 8{,}993 & 3{,}846 \\
    \bottomrule
\end{tabular}
}

\end{table*}


\section{Extended ablation studies}\label{sec:extended_ablation}

\begin{figure*}[t]
    \centering

    \begin{minipage}[c]{0.32\textwidth}
        \centering
        \includegraphics[width=\linewidth]{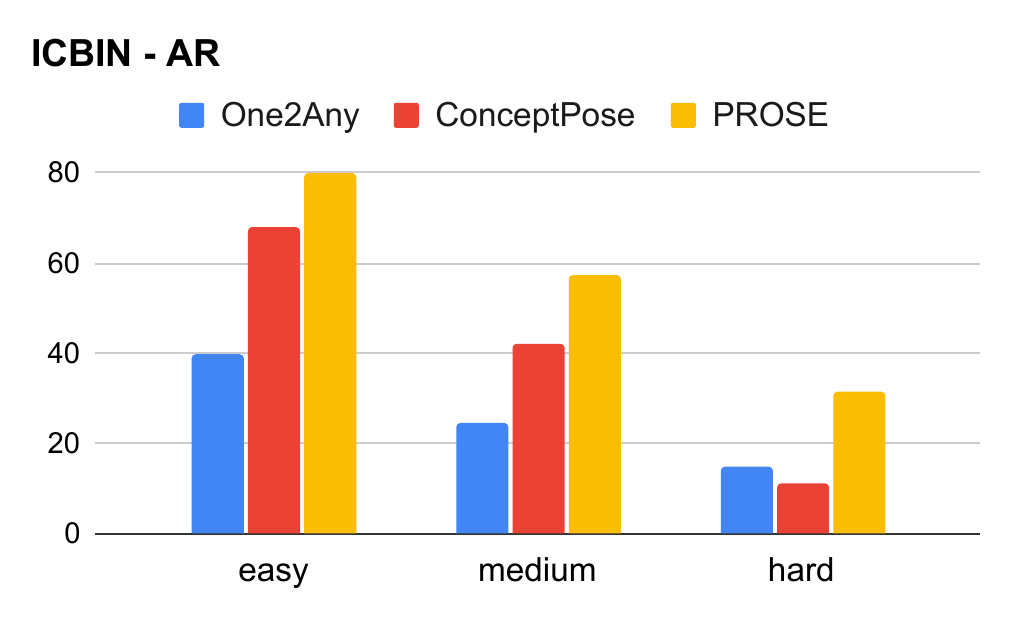}
    \end{minipage}
    \hfill
    \begin{minipage}[c]{0.32\textwidth}
        \centering
        \includegraphics[width=\linewidth]{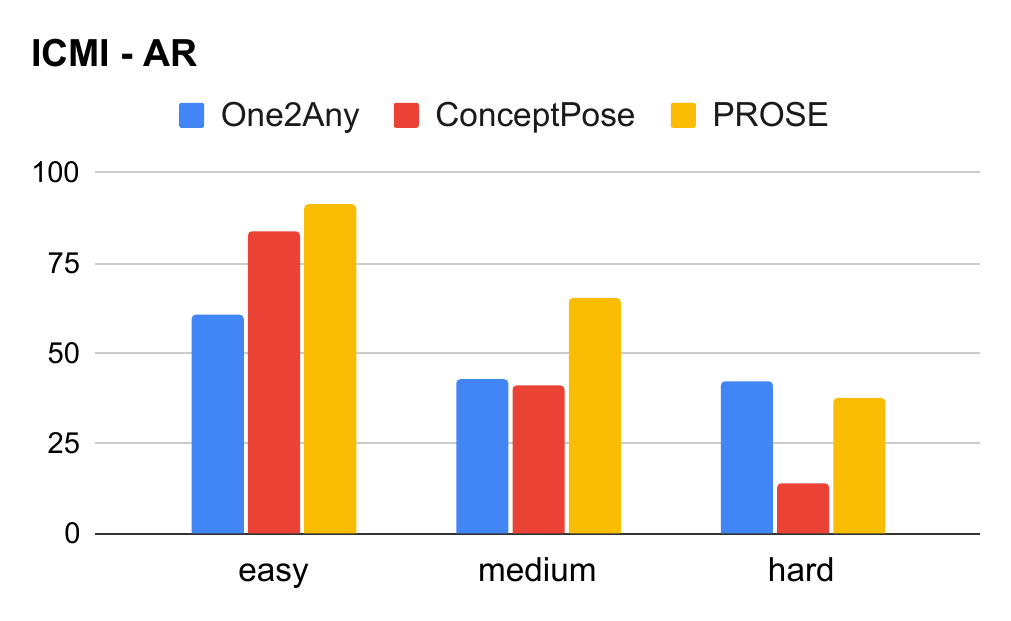}
    \end{minipage}
    \hfill
    \begin{minipage}[c]{0.32\textwidth}
        \centering
        \includegraphics[width=\linewidth]{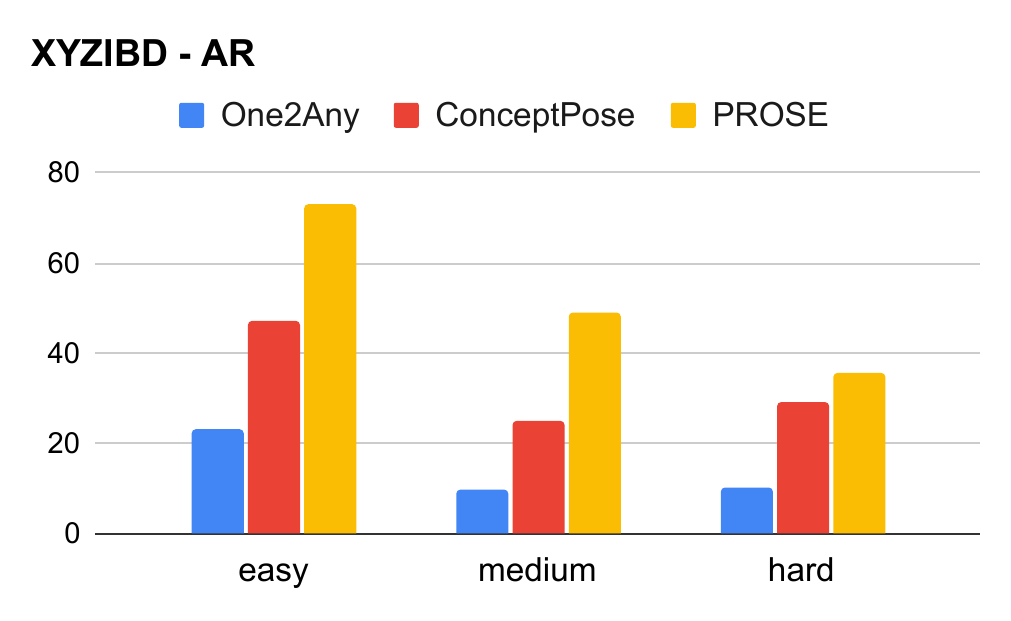}
    \end{minipage}

    \caption{Per-difficulty average recall of PROSE vs. the
    One2Any and ConceptPose baselines on IC-BIN, IC-MI, and XYZ-IBD, with
    instances split by anchor–instance overlap into easy (IoU\,$>$\,0.6), medium
    (0.3\,$<$\,IoU\,$<$\,0.6), and hard (0\,$<$\,IoU\,$<$\,0.3).}
    
    \label{fig:per_difficulty}
\end{figure*}

Figure~\ref{fig:per_difficulty} decomposes average recall by anchor–instance overlap,
splitting instances into easy (IoU\,$>$\,0.6), medium (0.3\,$<$\,IoU\,$<$\,0.6), and hard
(0\,$<$\,IoU\,$<$\,0.3). As expected, accuracy decreases with overlap for every method,
since a smaller visible surface overlap yields fewer reliable correspondences. PROSE
attains the highest AR in almost every bin across the three datasets, and its margin is
largest in the easy and medium regimes, \eg, on XYZ-IBD it reaches $72.9\%$/$48.8\%$
(easy/medium) against $47.0\%$/$25.0\%$ for ConceptPose and $23.0\%$/$10.0\%$ for One2Any.
The two baselines show no consistent ordering: ConceptPose is generally stronger on easy
instances, while One2Any is more competitive on the hardest bins of IC-BIN and IC-MI. The only setting where
PROSE is marginally surpassed is the hard split of IC-MI ($37.8\%$ vs.\ One2Any's
$42.0\%$), which comprises just $42$ low-overlap instances. Overall, PROSE's
gains are not limited to easy cases but persist across the full difficulty spectrum,
confirming its benefits in low-overlap scenarios. 

\begin{figure*}[!htbp]
    \centering
    \includegraphics[width=1.0\linewidth]{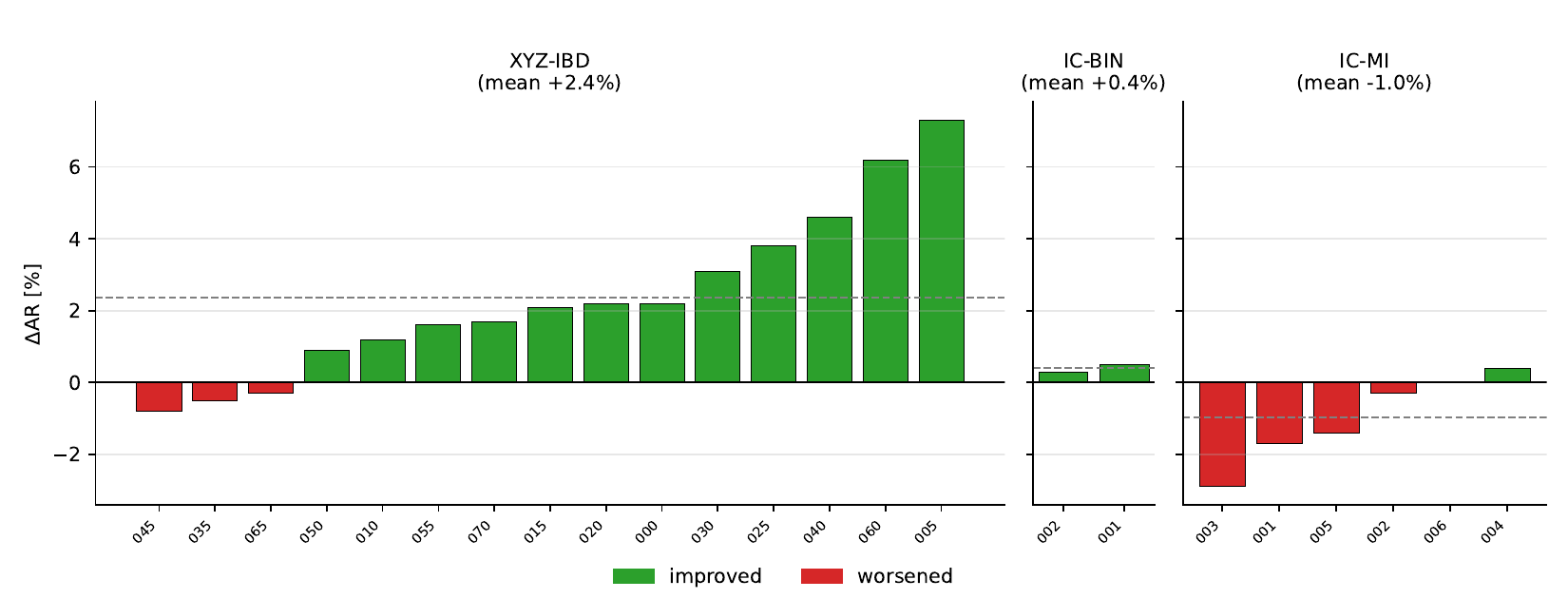}

    \caption{Change in average
    recall induced by the refinement, $\Delta\mathrm{AR}=\mathrm{AR}_{\text{after}}-
    \mathrm{AR}_{\text{before}}$, for every scene of XYZ-IBD, IC-BIN, and IC-MI (bars sorted
    within each dataset; green $=$ improved, red $=$ worsened; dashed line $=$ dataset mean).
    Refinement helps most on the many-instance XYZ-IBD scenes (mean $+2.4$), is
    neutral on IC-BIN ($+0.4$), and is slightly detrimental on the few-instance IC-MI
    ($-1.0$).}
    
    \label{fig:delta_bars}
\end{figure*}

Figure~\ref{fig:delta_bars} reports the change in average recall produced by the refinement, $\Delta\mathrm{AR}=\mathrm{AR}_{\text{after}}-
\mathrm{AR}_{\text{before}}$, for every scene of the three datasets (green: improved;
red: worsened; dashed line: dataset mean). The effect is strongly dataset-dependent and
tracks the number of co-visible instances. On XYZ-IBD, whose bins contain many instances
of the same industrial part, refinement improves $12$ of $15$ scenes for a mean gain of
$+2.4$~AR, reaching up to $+7$ on the densest scenes. On IC-BIN the effect is essentially
neutral (mean $+0.4$), whereas on IC-MI, whose scenes contain only about three
instances on average, refinement is marginally detrimental (mean $-1.0$), since too few
overlapping cycles are available for the synchronization to exploit. This confirms
the trend in Fig.~5 of the main paper: enforcing cycle consistency becomes beneficial
 when enough mutual references co-occur, and its gain grows with the number of
instances. Crucially, the many-instance regime is exactly the setting our method targets
(bin-picking of numerous identical objects, as in XYZ-IBD), so the refinement contributes
its largest improvements precisely where the task demands them.

\section{Additional quantitative results}\label{sec:additional_quant}

\newcolumntype{H}{>{\columncolor{myazure}}r}
\renewcommand{\arraystretch}{0.9}

\begin{table*}[t]
\centering
\tabcolsep 5pt
\caption{
Per-scene effect of refinement across the three datasets.
AR (BOP) and ADD-S in \%\ (higher better); RE\,[$^\circ$] and TE\,[mm]
(lower better).
All runs use identical settings, only the correspondences fed to RANSAC
differ (bef.: initial correspondences; aft.: refined).
Our refined results are \colorbox{myazure}{highlighted}.
Imp. denotes the absolute change, computed as aft.~$-$~bef.
Green indicates improvement or no degradation, while red indicates worse performance.
}
\label{tab:refine_perscene}
\vspace{-2mm}

\resizebox{\textwidth}{!}{%
\begin{tabular}{ll|rH r|rH r|rH r|rH r}
\toprule
& & \multicolumn{3}{c|}{AR (BOP) [\%] $\uparrow$}
  & \multicolumn{3}{c|}{ADD-S [\%] $\uparrow$}
  & \multicolumn{3}{c|}{RE [$^\circ$] $\downarrow$}
  & \multicolumn{3}{c}{TE [mm] $\downarrow$} \\
Dataset & Scene
& bef. & aft. & Imp.
& bef. & aft. & Imp.
& bef. & aft. & Imp.
& bef. & aft. & Imp. \\
\toprule

\multirow{2}{*}{IC-BIN}
 & 000001
 & 37.1 & \textbf{37.5} & \textcolor{mygreen}{+0.4}
 & 82.2 & \textbf{84.2} & \textcolor{mygreen}{+2.0}
 & 59.8 & \textbf{59.6} & \textcolor{mygreen}{-0.2}
 & 15.6 & \textbf{15.3} & \textcolor{mygreen}{-0.3} \\

 & 000002
 & 39.1 & \textbf{39.4} & \textcolor{mygreen}{+0.3}
 & \textbf{83.1} & 80.2 & \textcolor{myred}{-2.9}
 & 70.7 & \textbf{70.0} & \textcolor{mygreen}{-0.7}
 & \textbf{21.7} & 24.9 & \textcolor{myred}{+3.2} \\

\midrule

\multirow{6}{*}{IC-MI}
 & 000001
 & \textbf{64.7} & 63.0 & \textcolor{myred}{-1.7}
 & \textbf{100.0} & \textbf{100.0} & \textcolor{mygreen}{0.0}
 & \textbf{41.6} & 42.0 & \textcolor{myred}{+0.4}
 & \textbf{4.5} & 4.8 & \textcolor{myred}{+0.3} \\

 & 000002
 & \textbf{63.4} & 63.1 & \textcolor{myred}{-0.3}
 & \textbf{100.0} & \textbf{100.0} & \textcolor{mygreen}{0.0}
 & \textbf{46.3} & 46.8 & \textcolor{myred}{+0.5}
 & \textbf{2.4} & 2.6 & \textcolor{myred}{+0.2} \\

 & 000003
 & \textbf{87.2} & 84.3 & \textcolor{myred}{-2.9}
 & \textbf{100.0} & \textbf{100.0} & \textcolor{mygreen}{0.0}
 & \textbf{9.3} & 11.2 & \textcolor{myred}{+1.9}
 & \textbf{4.1} & \textbf{4.1} & \textcolor{mygreen}{0.0} \\

 & 000004
 & 74.3 & \textbf{74.7} & \textcolor{mygreen}{+0.4}
 & \textbf{100.0} & \textbf{100.0} & \textcolor{mygreen}{0.0}
 & \textbf{32.7} & \textbf{32.7} & \textcolor{mygreen}{0.0}
 & 3.9 & \textbf{3.6} & \textcolor{mygreen}{-0.3} \\

 & 000005
 & \textbf{73.8} & 72.4 & \textcolor{myred}{-1.4}
 & \textbf{100.0} & \textbf{100.0} & \textcolor{mygreen}{0.0}
 & 15.9 & \textbf{15.8} & \textcolor{mygreen}{-0.1}
 & \textbf{6.0} & 6.3 & \textcolor{myred}{+0.3} \\

 & 000006
 & \textbf{72.1} & \textbf{72.1} & \textcolor{mygreen}{0.0}
 & \textbf{100.0} & \textbf{100.0} & \textcolor{mygreen}{0.0}
 & 23.6 & \textbf{21.7} & \textcolor{mygreen}{-1.9}
 & \textbf{7.3} & 7.7 & \textcolor{myred}{+0.4} \\

\midrule

\multirow{15}{*}{XYZ-IBD}
 & 000000
 & 72.9 & \textbf{75.1} & \textcolor{mygreen}{+2.2}
 & 96.5 & \textbf{97.1} & \textcolor{mygreen}{+0.6}
 & 6.8 & \textbf{5.8} & \textcolor{mygreen}{-1.0}
 & 2.8 & \textbf{2.5} & \textcolor{mygreen}{-0.3} \\

 & 000005
 & 34.8 & \textbf{42.1} & \textcolor{mygreen}{+7.3}
 & 58.6 & \textbf{66.9} & \textcolor{mygreen}{+8.3}
 & 51.8 & \textbf{43.2} & \textcolor{mygreen}{-8.6}
 & 23.0 & \textbf{15.8} & \textcolor{mygreen}{-7.2} \\

 & 000010
 & 61.1 & \textbf{62.3} & \textcolor{mygreen}{+1.2}
 & 98.3 & \textbf{98.4} & \textcolor{mygreen}{+0.1}
 & 9.0 & \textbf{8.1} & \textcolor{mygreen}{-0.9}
 & 4.9 & \textbf{4.7} & \textcolor{mygreen}{-0.2} \\

 & 000015
 & 63.5 & \textbf{65.6} & \textcolor{mygreen}{+2.1}
 & 93.9 & \textbf{94.7} & \textcolor{mygreen}{+0.8}
 & 22.0 & \textbf{19.4} & \textcolor{mygreen}{-2.6}
 & 7.9 & \textbf{7.2} & \textcolor{mygreen}{-0.7} \\

 & 000020
 & 46.0 & \textbf{48.2} & \textcolor{mygreen}{+2.2}
 & 84.9 & \textbf{86.7} & \textcolor{mygreen}{+1.8}
 & 24.3 & \textbf{22.3} & \textcolor{mygreen}{-2.0}
 & 9.0 & \textbf{8.0} & \textcolor{mygreen}{-1.0} \\

 & 000025
 & 61.9 & \textbf{65.7} & \textcolor{mygreen}{+3.8}
 & 95.6 & \textbf{96.2} & \textcolor{mygreen}{+0.6}
 & 12.5 & \textbf{6.7} & \textcolor{mygreen}{-5.8}
 & 6.2 & \textbf{5.4} & \textcolor{mygreen}{-0.8} \\

 & 000030
 & 44.5 & \textbf{47.6} & \textcolor{mygreen}{+3.1}
 & 87.7 & \textbf{89.8} & \textcolor{mygreen}{+2.1}
 & 36.8 & \textbf{33.4} & \textcolor{mygreen}{-3.4}
 & 10.7 & \textbf{8.9} & \textcolor{mygreen}{-1.8} \\

 & 000035
 & \textbf{75.3} & 74.8 & \textcolor{myred}{-0.5}
 & \textbf{100.0} & \textbf{100.0} & \textcolor{mygreen}{0.0}
 & 3.8 & \textbf{3.6} & \textcolor{mygreen}{-0.2}
 & 6.6 & \textbf{6.4} & \textcolor{mygreen}{-0.2} \\

 & 000040
 & 62.4 & \textbf{67.0} & \textcolor{mygreen}{+4.6}
 & 98.7 & \textbf{99.2} & \textcolor{mygreen}{+0.5}
 & 19.3 & \textbf{12.0} & \textcolor{mygreen}{-7.3}
 & 6.1 & \textbf{5.3} & \textcolor{mygreen}{-0.8} \\

 & 000045
 & \textbf{38.9} & 38.1 & \textcolor{myred}{-0.8}
 & \textbf{92.5} & 92.1 & \textcolor{myred}{-0.4}
 & \textbf{69.9} & \textbf{69.9} & \textcolor{mygreen}{0.0}
 & 10.7 & \textbf{10.5} & \textcolor{mygreen}{-0.2} \\

 & 000050
 & 42.2 & \textbf{43.1} & \textcolor{mygreen}{+0.9}
 & 93.3 & \textbf{93.7} & \textcolor{mygreen}{+0.4}
 & 53.6 & \textbf{51.1} & \textcolor{mygreen}{-2.5}
 & 16.3 & \textbf{15.0} & \textcolor{mygreen}{-1.3} \\

 & 000055
 & 56.0 & \textbf{57.6} & \textcolor{mygreen}{+1.6}
 & \textbf{100.0} & \textbf{100.0} & \textcolor{mygreen}{0.0}
 & 74.2 & \textbf{69.4} & \textcolor{mygreen}{-4.8}
 & 32.8 & \textbf{24.5} & \textcolor{mygreen}{-8.3} \\

 & 000060
 & 50.8 & \textbf{57.0} & \textcolor{mygreen}{+6.2}
 & 88.4 & \textbf{92.1} & \textcolor{mygreen}{+3.7}
 & 24.1 & \textbf{15.7} & \textcolor{mygreen}{-8.4}
 & 12.7 & \textbf{10.6} & \textcolor{mygreen}{-2.1} \\

 & 000065
 & \textbf{64.6} & 64.3 & \textcolor{myred}{-0.3}
 & 97.8 & \textbf{98.0} & \textcolor{mygreen}{+0.2}
 & \textbf{28.3} & 30.2 & \textcolor{myred}{+1.9}
 & 23.0 & \textbf{21.9} & \textcolor{mygreen}{-1.1} \\

 & 000070
 & 30.5 & \textbf{32.1} & \textcolor{mygreen}{+1.6}
 & 82.0 & \textbf{84.8} & \textcolor{mygreen}{+2.8}
 & 87.8 & \textbf{85.5} & \textcolor{mygreen}{-2.3}
 & 13.7 & \textbf{12.6} & \textcolor{mygreen}{-1.1} \\

\bottomrule
\end{tabular}
}
\end{table*}

Table~\ref{tab:refine_perscene} isolates the contribution of the
refinement: for each scene we report the four metrics obtained when RANSAC is fed the
initial nearest-neighbour correspondences (\emph{bef.}) versus the refined ones
(\emph{aft.}), holding all other components fixed. On XYZ-IBD, refinement improves AR on $12$ of the $15$ scenes and reduces both
rotation and translation error on almost all of them; the largest AR gains are $+7.3$
(scene 000005, $51.8^\circ\!\rightarrow\!43.2^\circ$ RE, $23.0\!\rightarrow\!15.8$~mm TE),
$+6.2$ (000060, $24.1^\circ\!\rightarrow\!15.7^\circ$ RE) and $+4.6$ (000040,
$19.3^\circ\!\rightarrow\!12.0^\circ$ RE). On the remaining three scenes (000035, 000045,
000065) AR changes by at most $0.8$ points, \ie\ refinement leaves them essentially
unchanged. On IC-BIN the effect is small but positive (AR and RE improve on both scenes),
while on IC-MI it is neutral to slightly negative: ADD-S is already saturated at $100\%$
and the other metrics move by at most a few points (largest drop $-2.9$~AR on 000003).
This dataset-level pattern is consistent with the per-scene $\Delta$AR summary
of Fig.~\ref{fig:delta_bars}
 and the instance-count trend of Fig.~5: refinement helps
most where many instances co-occur (XYZ-IBD) and little where only a few are available
(IC-MI, $\sim$3 per scene). Overall, refinement never causes a large regression (worst
per-scene AR drop $2.9$ points) while yielding consistent gains in the many-instance
regime that motivates our setting.

Table~\ref{tab:before_after_refine_400_fps_aggregate} reports the dataset-level effect of
cycle-consistency refinement comparing the metrics obtained when RANSAC is fed the
initial nearest-neighbour correspondences (\emph{before}) versus the refined ones
(\emph{after}), with all other components held fixed. On XYZ-IBD refinement improves \emph{every} metric: AR from
$53.2$ to $55.8$ ($+2.6$ points), ADD-S from $90.6$ to $92.1$, rotation error from
$32.1^\circ$ to $28.6^\circ$ ($-3.5^\circ$), and translation error from $10.6$ to
$9.1$~mm ($-1.5$~mm), a substantial and consistent gain. On IC-BIN the effect is only
marginal (AR $+0.4$, ADD-S $+0.9$, RE $-0.3^\circ$), with translation error slightly
worse ($+0.5$~mm). On IC-MI refinement is marginally detrimental (AR $-0.9$,
RE $+0.1^\circ$, TE $+0.2$~mm), while ADD-S is already saturated at $100\%$. This
ordering (clear gains on XYZ-IBD, near-neutral on IC-BIN, slightly negative on
IC-MI) matches the per-scene breakdown (Table~\ref{tab:refine_perscene},
Fig.~\ref{fig:delta_bars}) and follows the instance-count trend of Fig.~5:
synchronization exploits redundancy across co-visible instances, so its benefit scales
with the number of instances per view ($26.5$ on XYZ-IBD versus $2.7$ on IC-MI).

\newcolumntype{H}{>{\columncolor{myazure}}r}
\renewcommand{\arraystretch}{0.9}

\begin{table*}[t]
\centering
\tabcolsep 6pt
\caption{
Effect of global refinement on relative 6D-pose accuracy,
aggregated per dataset.
AR (BOP) and ADD-S are reported in \%\ (higher better);
RE\,[$^\circ$] and TE\,[mm] are lower better.
All runs use identical settings, only the correspondences fed to RANSAC
differ (before: initial correspondences; after: refined).
Our refined results are \colorbox{myazure}{highlighted}.
Imp. denotes the absolute change, computed as after~$-$~before.
Green indicates improvement or no degradation, while red indicates worse performance.
}
\label{tab:before_after_refine_400_fps_aggregate}
\vspace{-2mm}

\resizebox{\textwidth}{!}{%
\begin{tabular}{l|rHr|rHr|rHr|rHr}
\toprule
& \multicolumn{3}{c|}{AR [\%] (BOP) $\uparrow$}
& \multicolumn{3}{c|}{ADD-S [\%] $\uparrow$}
& \multicolumn{3}{c|}{RE\,[$^\circ$] $\downarrow$}
& \multicolumn{3}{c}{TE\,[mm] $\downarrow$} \\

Dataset
& before & after & Imp.
& before & after & Imp.
& before & after & Imp.
& before & after & Imp. \\
\toprule

XYZ-IBD
& 53.2 & \textbf{55.8} & \textcolor{mygreen}{+2.6}
& 90.6 & \textbf{92.1} & \textcolor{mygreen}{+1.5}
& 32.13 & \textbf{28.63} & \textcolor{mygreen}{-3.50}
& 10.62 & \textbf{9.09} & \textcolor{mygreen}{-1.53} \\

IC-BIN
& 37.6 & \textbf{38.0} & \textcolor{mygreen}{+0.4}
& 82.4 & \textbf{83.3} & \textcolor{mygreen}{+0.9}
& 62.36 & \textbf{62.03} & \textcolor{mygreen}{-0.33}
& \textbf{17.01} & 17.54 & \textcolor{myred}{+0.53} \\

IC-MI
& \textbf{71.2} & 70.3 & \textcolor{myred}{-0.9}
& \textbf{100.0} & \textbf{100.0} & \textcolor{mygreen}{0.0}
& \textbf{30.60} & 30.74 & \textcolor{myred}{+0.14}
& \textbf{4.49} & 4.64 & \textcolor{myred}{+0.15} \\

\bottomrule
\end{tabular}
}
\end{table*}
\section{Additional qualitative results}\label{sec:additional_qual}

\begin{figure}[t!]
    \centering
    \includegraphics[width=1.0\linewidth]{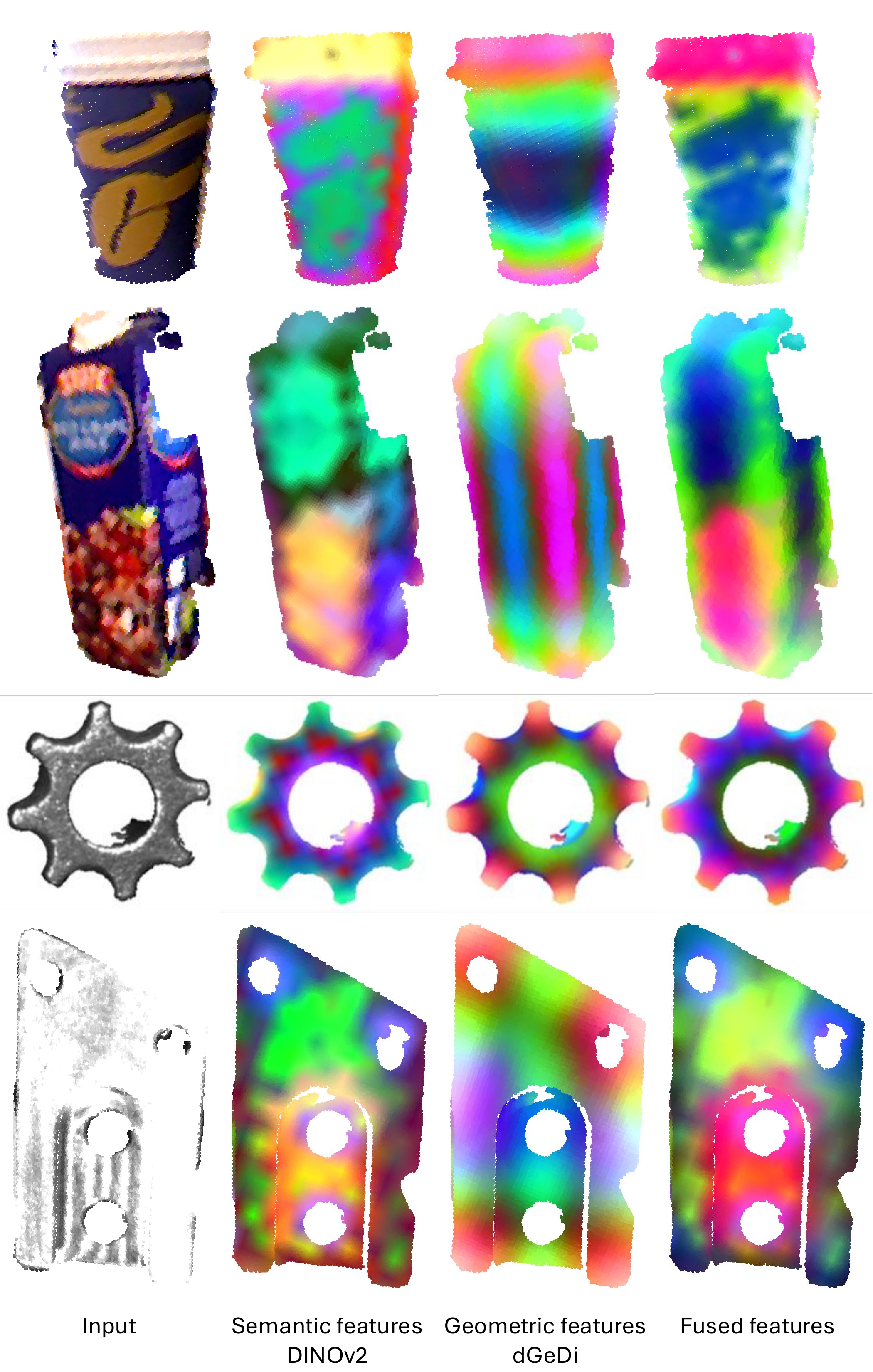}

    \caption{
    Qualitative comparison of appearance-aware and semantic-aware, geometry-aware, and fused feature representations for different object categories.
    From top to bottom: coffee cups, juice boxes, gears, and metal brackets.
    The left column shows the input object crop, the second column shows DINOv2  features, the third column shows dGeDi features, and the fourth column shows fused features. Here we used Principal Component Analysis (PCA) to reduce the original feature size to 3 for visualization purpose. DINOv2 captures stronger semantic and appearance cues, which are particularly useful for textured objects such as coffee cups and juice boxes. In contrast, dGeDi captures geometric structure more effectively, making it better suited for texture-less industrial objects such as gears and metal brackets.
    } 
    \label{fig:instance_features}
\end{figure}

\begin{figure}[t!]
    \centering
    \includegraphics[width=1.0\linewidth]{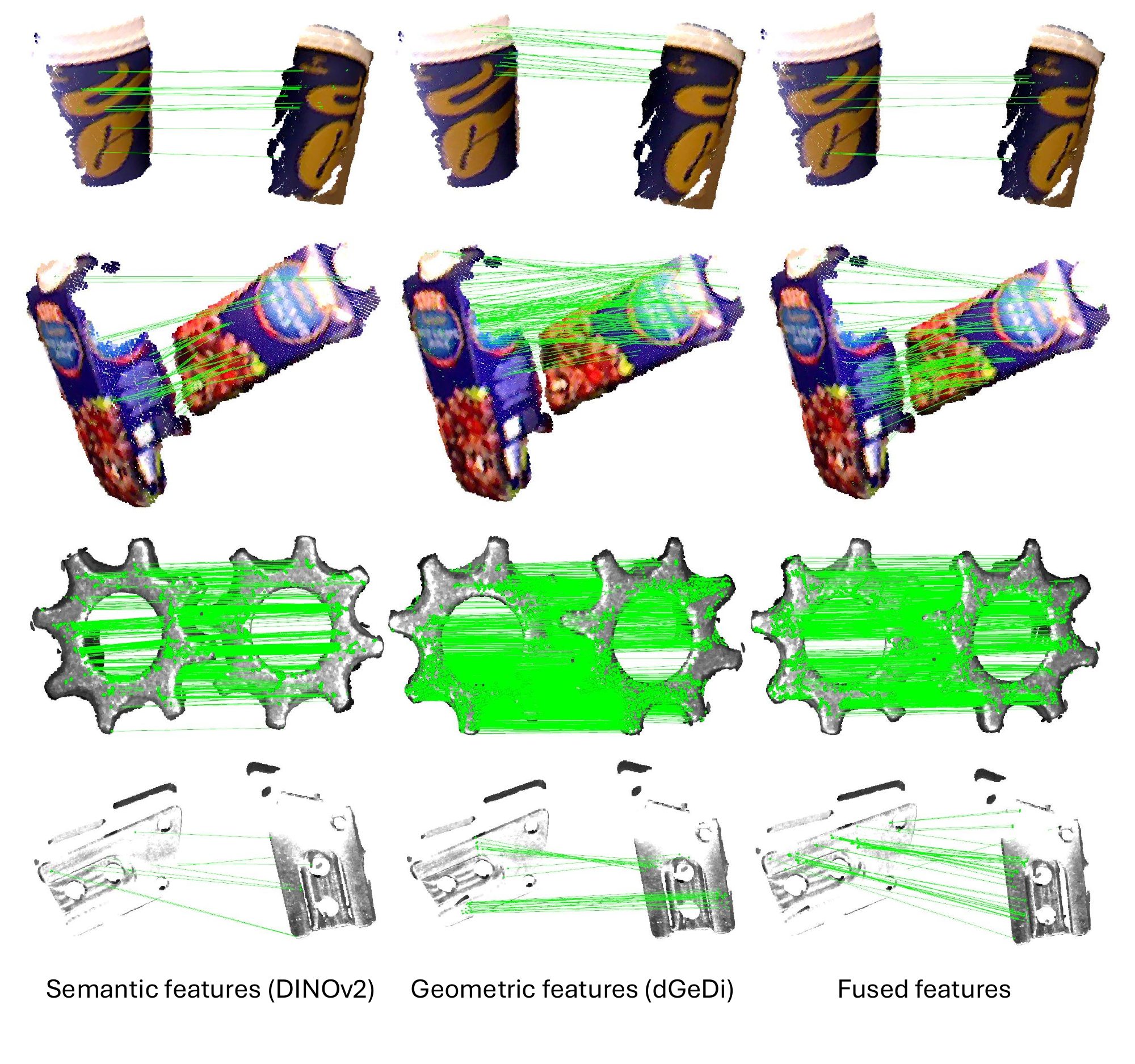}

    \caption{
    Qualitative comparison of inlier correspondences obtained using
    semantic (DINOv2), geometric (dGeDi), and fused feature representations.
    From top to bottom, we show coffee cups, juice boxes, gears, and metal
    brackets. Semantic features provide stronger appearance cues for
    textured objects, whereas geometric features are more informative for
    textureless industrial objects. The fused representation combines these
    complementary cues and generally produces a denser and more robust set
    of inlier correspondences across object types.
    }
    \label{fig:inlier_corrs}
\end{figure}

Figure~\ref{fig:instance_features} visualizes the learned descriptors used by our method,
reduced to three dimensions with PCA, for four representative objects that
span textured everyday items (coffee cups, juice boxes) and texture-less industrial parts
(gears, metal brackets). Two observations motivate the fused representation. First, within
each object the colours are stable across corresponding surface regions, indicating that the
descriptors are repeatable and discriminative: exactly the property our coarse matching
relies on. Second, the two modalities are complementary: DINOv2 produces semantic features on textured objects (\eg\ it separates the label, body, and rim of a cup) but
collapses to meaningless representations on the untextured gears and brackets, where appearance
carries little information; dGeDi behaves oppositely, encoding local geometry (gear teeth,
bracket edges and holes) and thus remaining discriminative precisely where DINOv2 fails,
while being less informative on smooth, textured surfaces. The fused descriptor, obtained by
concatenating the two, inherits the strengths of both and stays
discriminative across all four categories. This explains why fusion attains the best average
recall in our feature ablation (Table 2 of the main paper) without any per-object
tuning, and why we adopt it as the default: a single representation that transfers from
everyday textured objects to texture-less industrial parts.

Figure~\ref{fig:inlier_corrs} shows how the complementary properties observed in the descriptor visualizations translate into the correspondences used for pose estimation. For textured everyday objects, such as coffee cups and juice boxes, DINOv2 produces reliable matches between visually distinctive regions, whereas its correspondences become less discriminative on textureless industrial objects. Conversely, dGeDi yields stronger correspondences on gears and metal brackets by exploiting distinctive geometric structures, such as teeth, edges, and holes, but provides fewer informative matches on smooth surfaces. By combining these cues, the fused representation preserves appearance-based matches where texture is informative and introduces geometry-based matches where appearance is ambiguous. Consequently, it generally produces a denser and more spatially distributed set of inlier correspondences across object categories. Together with Figure~\ref{fig:instance_features}, these results show that fusion improves both descriptor discriminability and its practical outcome during coarse matching, providing a robust representation across the heterogeneous objects considered in our benchmark.

\begin{figure*}[t]
    \centering

    \begin{minipage}[t]{0.48\textwidth}
        \centering
        \includegraphics[width=\linewidth]{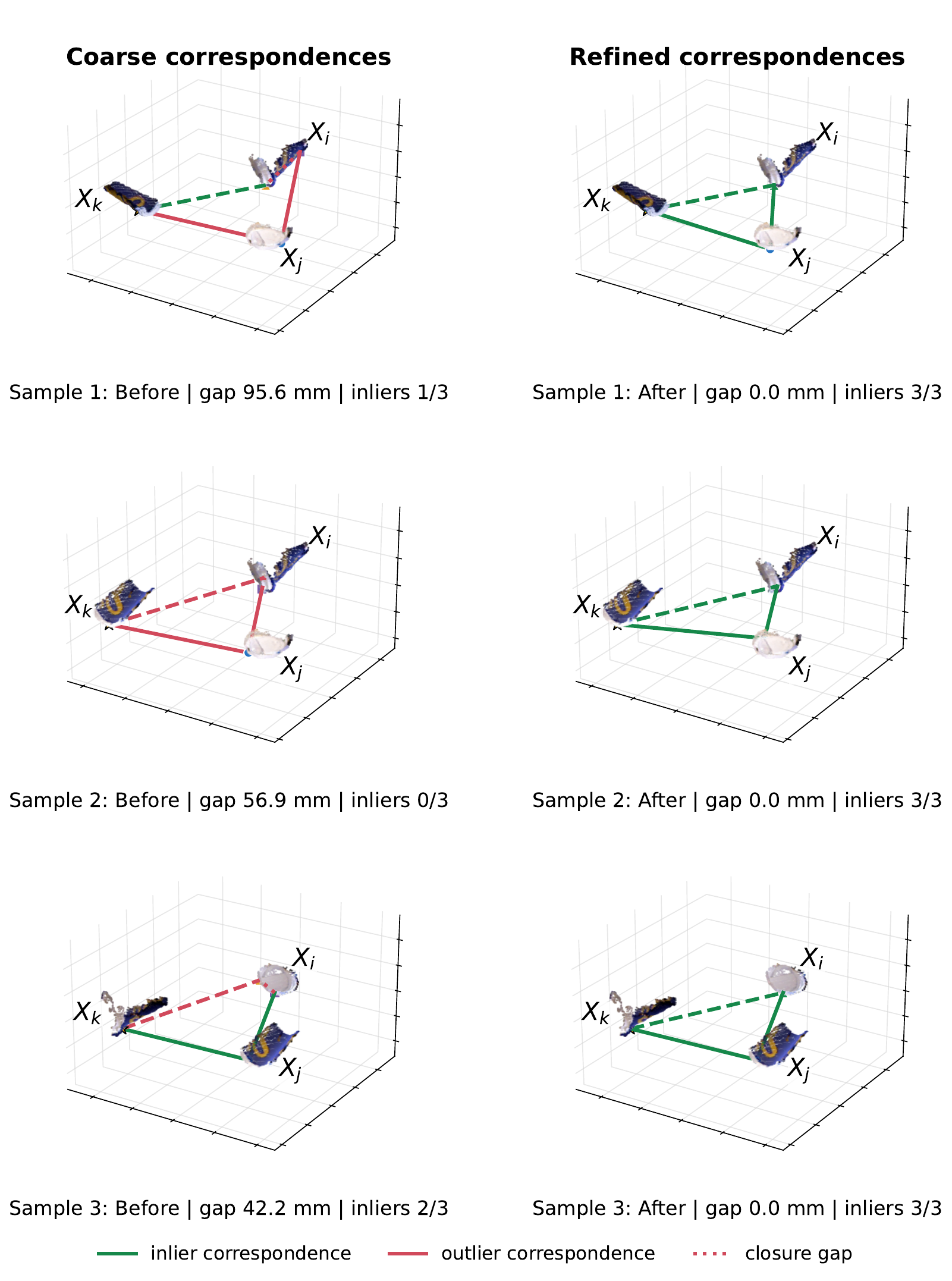}
        
        \vspace{2pt}
        (a) Success cases
    \end{minipage}
    \hfill
    \begin{minipage}[t]{0.48\textwidth}
        \centering
        \includegraphics[width=\linewidth]{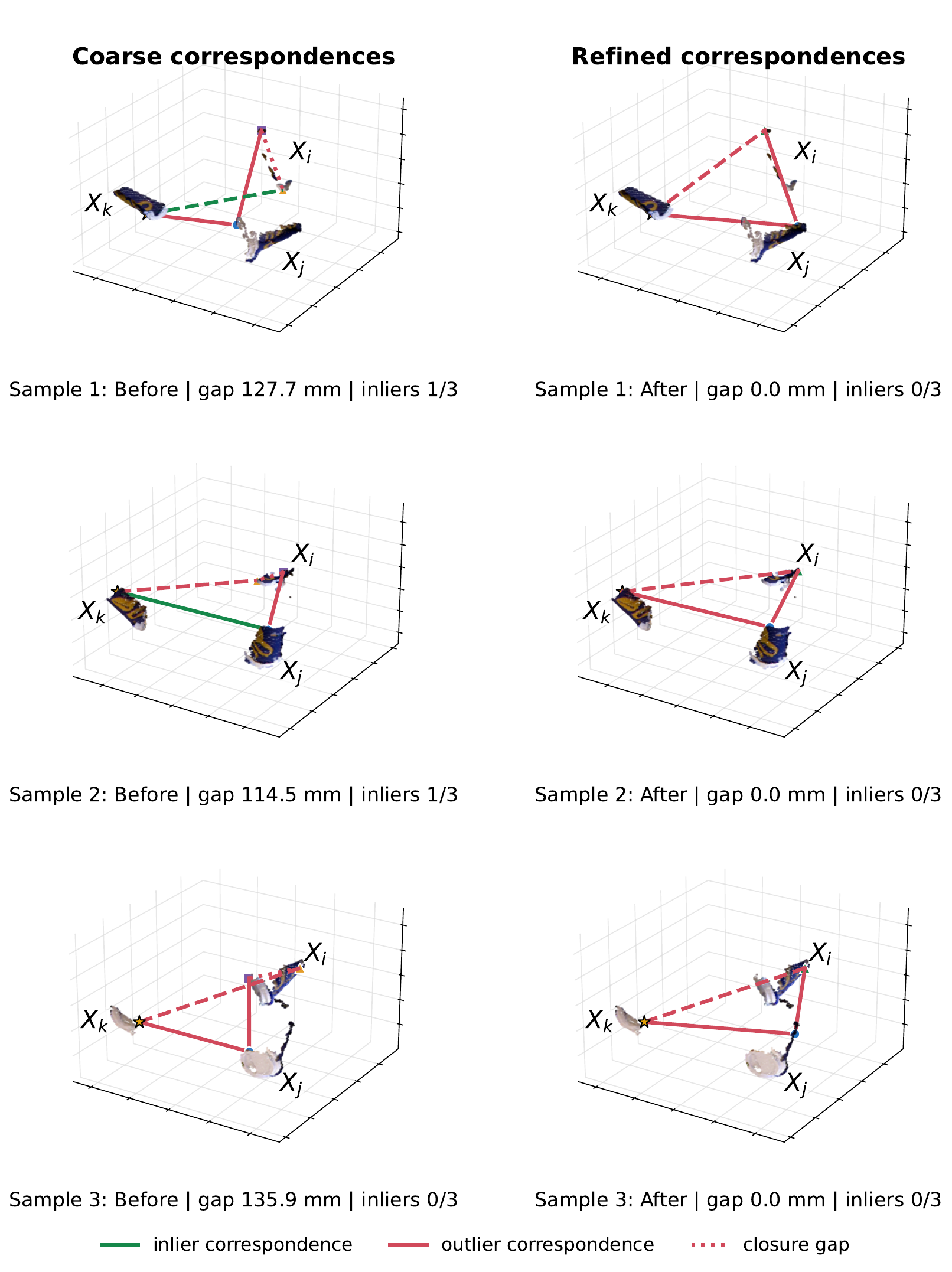}
        
        \vspace{2pt}
        (b) Failure cases
    \end{minipage}

    \caption{Cycle consistency: successes and failures. Each row is one instance
    triplet (anchor $X_k$, mediator $X_j$, target $X_i$); the \emph{left} column shows the
    coarse (independent nearest-neighbour) correspondences and the \emph{right} column the
    correspondences after cycle-consistency refinement. Green/red lines are symmetry-aware
    inlier/outlier correspondences and the dashed red line is the cycle-closure gap.
    (a) Successes: the coarse matches are inconsistent (closure gaps of $42$--$96$\,mm
    and only $0$--$2$ of $3$ inliers), and refinement closes the cycle (gap $\rightarrow 0$)
    while recovering all three correct edges ($3/3$ inliers). (b) Failures: refinement
    still closes the cycle (gap $\rightarrow 0$) but converges to a geometrically wrong solution
    ($0/3$ inliers); in the first two rows it even discards the single correct coarse edge
    ($1/3\rightarrow0/3$), illustrating that cycle consistency enforces mutual agreement, not
    ground-truth correctness.}
    \label{fig:success_failure}
\end{figure*}

Figure~\ref{fig:success_failure} visualizes how the refinement step acts on individual three-instance
cycles. Each panel draws, for a triplet (anchor $X_k$, mediator $X_j$, target $X_i$), the
two composed correspondences and the direct one; under perfect cycle consistency the
composed path and the direct correspondence coincide, so the dashed \emph{closure gap}
vanishes. In the success cases~(a), the coarse per-pair matches disagree (closure gaps of
$42$--$96$\,mm and only $0$--$2$ of $3$ geometric inliers) and the synchronization resolves
them into a fully consistent and correct cycle ($3/3$ inliers, $0$\,mm gap). This is
the intended behaviour and the source of the accuracy gains reported in the main paper. The
failure cases~(b) expose the limitation of the approach: refinement always yields a
\emph{consistent} cycle (gap $\rightarrow 0$ by construction), but consistency does not imply
correctness. When the coarse descriptors are ambiguous, \eg\ repetitive, the low-rank synchronization can lock all three edges onto a
mutually agreeing yet wrong assignment ($0/3$ inliers), and may even overturn a correct
coarse edge to reach it (rows~1--2, $1/3\rightarrow0/3$). Such cases are infrequent in
aggregate (cf.\ Table~\ref{tab:refine_perscene}) and are further mitigated downstream by
robust RANSAC, but they constitute the principal failure mode of correspondence-level cycle
consistency and motivate coupling it with geometric verification.

\end{document}